\documentclass[10pt,twocolumn,letterpaper]{article}

\usepackage{cvpr}              
\definecolor{cvprblue}{rgb}{0.21,0.49,0.74}
\usepackage[pagebackref,breaklinks,colorlinks,allcolors=cvprblue]{hyperref}

\def\paperID{1405} 
\def\confName{CVPR}
\def\confYear{2026}

\title{Edit2Perceive: Image Editing Diffusion Models Are Strong Dense Perceivers}

\author{
  Yiqing Shi\textsuperscript{1}\textsuperscript{$\ast$} \quad
  Yiren Song\textsuperscript{2}\textsuperscript{$\ast$} \quad
  Mike Zheng Shou\textsuperscript{2}\textsuperscript{$\dagger$} \\
  \textsuperscript{1}Peking University, \textsuperscript{2}Show Lab, National University of Singapore \\
}

\begin{document}

\twocolumn[{%
\renewcommand\twocolumn[1][]{#1}%
\maketitle

\begin{center}
   \captionsetup{type=figure} 
    \includegraphics[height=6cm]{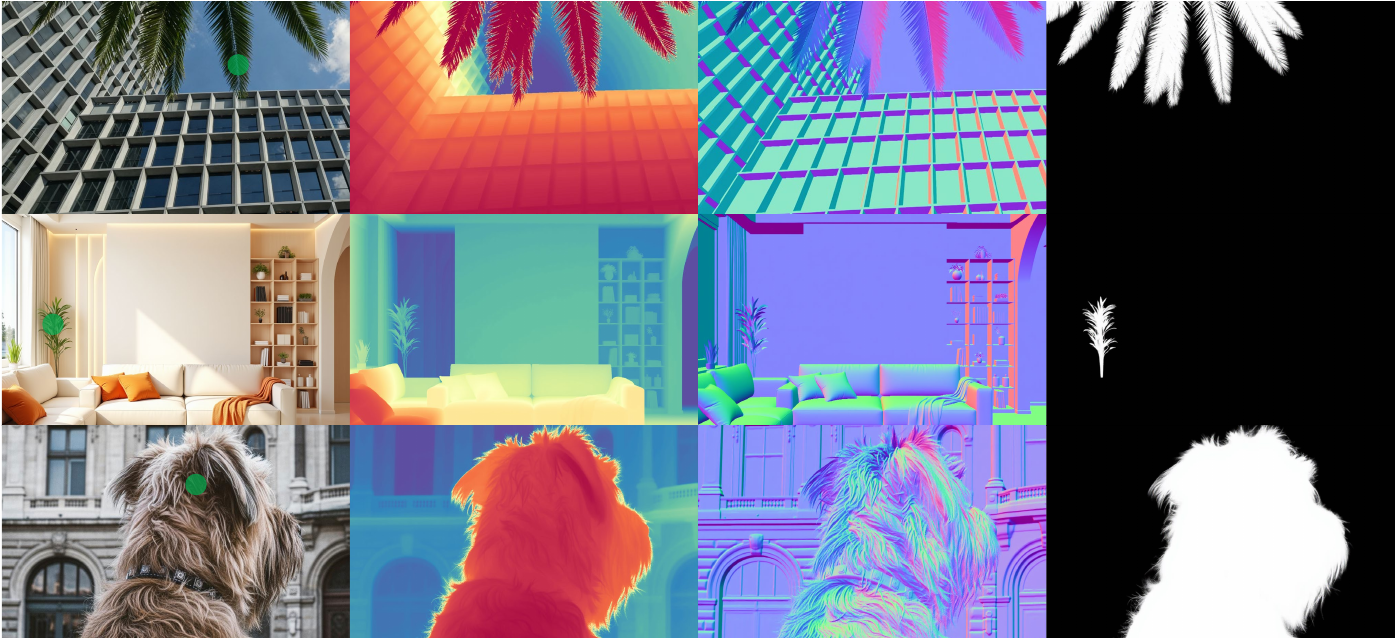} %
    \includegraphics[height=6cm]{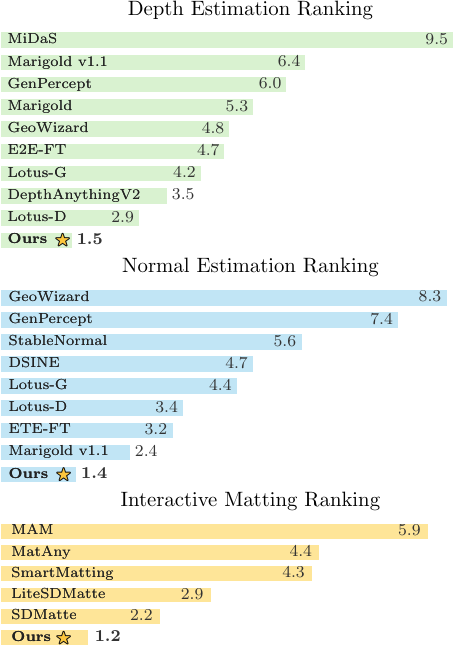}
    \setlength{\abovecaptionskip}{0pt} 
    \caption{We present \textbf{Edit2Perceive}, a unified framework for diverse dense prediction tasks. 
Our model achieves state-of-the-art performance across zero-shot Monocular Depth Estimation, Surface Normal Estimation, and Interactive Matting, consistently outperforming previous methods. The bar charts on the right compare the average ranking score of our method against prior work on each task; lower is better. The green dot (\textcolor{green}{$\bullet$}) on the input images indicates the positive user prompt for interactive matting.
}
    \label{teaser} 
    \vspace{5pt}
\end{center}   
}]

\begingroup
\renewcommand\thefootnote{}
\footnotetext{* Equal contribution.}
\footnotetext{$\dagger$ Corresponding author.}
\endgroup

\begin{abstract}
Recent advances in diffusion transformers have shown remarkable generalization in visual synthesis, yet most dense perception methods still rely on text-to-image (T2I) generators designed for stochastic generation. 
We revisit this paradigm and show that image editing diffusion models are inherently image-to-image consistent, providing a more suitable foundation for dense perception task. 
We introduce \textbf{Edit2Perceive}, a unified diffusion framework that adapts editing models for depth, normal, and matting. 
Built upon the FLUX.1 Kontext architecture, our approach employs full-parameter fine-tuning and a pixel-space consistency loss to enforce structure-preserving refinement across intermediate denoising states. 
Moreover, our single-step deterministic inference yields up to faster runtime while training on relatively small datasets.
Extensive experiments demonstrate comprehensive state-of-the-art results across all three tasks, revealing the strong potential of editing-oriented diffusion transformers for geometry-aware perception. Code is released at \href{https://github.com/showlab/Edit2Perceive}{https://github.com/showlab/Edit2Perceive}.
\end{abstract}


\section{Introduction}

\label{sec:intro}

Dense perception tasks, such as Monocular Depth Estimation, Surface Normal Estimation, and Interactive Matting, are central to understanding 3D structure and object segmentation from 2D images. These tasks are particularly challenging because they require models to accurately predict pixel-level geometric or optical attributes. Monocular Depth Estimation, for instance, necessitates recovering 3D scene geometry from a single image, which is inherently ill-posed. Similarly, Surface Normal Estimation demands precise angular predictions, while Interactive Matting involves the segmenting the foreground from the background in an image given interactive input, which is highly sensitive to edge details. Solving these ill-posed problems depends on how much prior knowledge a model possesses about the visual world.
Recent progress has been largely driven by leveraging rich visual priors from large-scale text-to-image (T2I) diffusion models, such as Stable Diffusion~\cite{SD}, which have shown impressive success when adapted for perception tasks~\cite{marigold}.

However, we argue that this paradigm is inherently limited by a representation mismatch.
T2I models are trained to synthesize diverse visual concepts based on unstructured textual prompts, which excel at semantic composition (``concept-to-pixel” mapping) but lack the capacity to reason about the structural relationships within an image.
As a result, they are misaligned with the deterministic and geometry-aware nature of dense perception.

We revisit this assumption and propose that image-to-image (I2I) diffusion models, especially context-based editors like FLUX.1 Kontext~\cite{flux-kontext}, offer a more natural foundation for dense perception.
Unlike T2I models, these models are trained to generate semantically coherent edits conditioned on an existing image. This pretraining objective implicitly requires parsing the input into a structured scene representation—capturing objects, surfaces, and their interrelations. Such structured priors is essential for dense perception.

Building upon this insight, we introduce \textbf{Edit2Perceive}, a unified framework that adapts a powerful I2I diffusion model into a dense perceiver. We first convert the stochastic denoising process into a pseudo-deterministic path by fixing the random seed, ensuring a unique and reproducible input-output mapping.To further strengthen geometric fidelity, we design a pixel-space consistency loss that enforces fine-grained geometric fidelity and enhances structure preservation in the final output. Additionally, to meet the high numerical stability required for Monocular Depth Estimation task, we theoretically derive that a square-root mapping minimizes the Absolute Reletive (AbsRel) error caused by the preprocess. Finally, benefiting from the inherent mechanism of flow matching, Edit2Perceive enables single-step deterministic inference, combining accuracy with efficiency.

Extensive experiments across three major dense perception tasks (Monocular Depth Estimation, Surface Normal Estimation, and Interactive Matting) show that our method achieves state-of-the-art performance while using limited train data. It achieved impressive results in the wild (Fig.~\ref{teaser}).

Our main contributions are summarized as follows:
\begin{itemize}
    \item We demonstrate that image editing diffusion models, rather than text-to-image generators, provide a better inductive bias for deterministic dense perception, effectively bridging generative modeling and geometric reasoning.
    \item We propose \textbf{Edit2Perceive}, a unified diffusion transformer fine-tuned across depth, normal, and matting tasks, enhanced with a pixel-consistency loss and an analytically derived optimal normalization for depth.
    \item We achieve state-of-the-art performance with efficient single-step inference, showing that editing-oriented diffusion models can serve as a new class of perception-oriented foundation models.
\end{itemize}

\begin{figure*}[!h]
\centering
\includegraphics[scale=0.52]{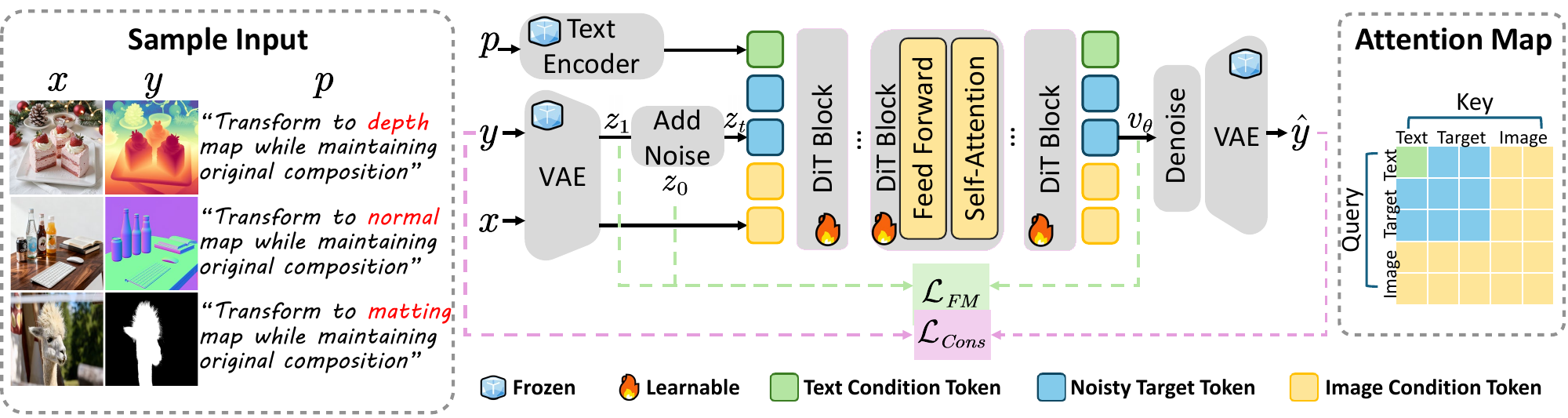}
\caption{\textbf{Overview of the Editor2Perceive Framework.}
We adapt the FLUX.1 Kontext editor for dense perception. 
An input image $x$ and text prompt $p$ condition for the target image $y$. In the forward process, target token will be adding noise $z_t$ and then concat with text condition token and image condition token. The DiT backbone is trained to predict the velocity of the flow from a noise vector $z_0$ to the target latent $z_1$. Our training objective combines the latent-space flow matching loss ($\mathcal{L}_{\text{FM}}$) with a pixel-space consistency loss ($\mathcal{L}_{\text{Cons}}$) for enhanced geometric fidelity. }
\label{fig:framework}
\end{figure*}

\section{Related Work}


\subsection{Image Generation and Editing Models}
Large-scale diffusion models have become the cornerstone of visual content creation. Text-to-image (T2I) systems such as Stable Diffusion~\cite{SD} and FLUX.1~\cite{flux} demonstrate strong semantic understanding through large-scale image--text pre-training. Meanwhile, research attention is shifting toward more controllable generation \cite{ma2024followyouremoji, ma2025followfaster, ma2025followyourclick, ma2025followyourmotion, song2024processpainter, zhang2024stable, zhang2025stable, chen2025transanimate}. Unlike T2I models that generate scenes from abstract text, I2I models aim to make precise modifications to existing visual content. Recent editing systems such as Step1X-Edit~\cite{step1-edit} and FLUX.1 Kontext~\cite{flux-kontext} learn complex correspondences between image structure and editing instructions, achieving high structural consistency. 
With the transition to Diffusion Transformers~\cite{dit}, controllable generation has advanced rapidly. Models such as EasyControl~\cite{zhang2025easycontrol} exemplify MM-DiT–based conditioning and have inspired a broad line of DiT-driven methods~\cite{song2025omniconsistency, gong2025relationadapter, jiang2025personalized, wang2025diffdecompose, song2025makeanything, song2025layertracer, guo2025any2anytryon,  lu2025easytext, shi2025wordcon, shi2024fonts}.

\subsection{Dense Perception Tasks}
Dense perception tasks, including monocular depth estimation, surface normal estimation, and interactive matting, are fundamental to computer vision.
Traditional approaches can be broadly categorized into two paths.
The first is data-driven, involving training specialized models on massive annotated datasets.
From MiDaS~\cite{midas}, which pioneered cross-dataset generalization, to the Depth Anything series~\cite{depth_anything_v1, depth_anything_v2}, which leverages large-scale unlabeled data for pre-training, this path improves performance by scaling up data and model size.
However, this approach is computationally intensive, and the resulting models are often designed for a single task, limiting their ability to generalize to new ones.
The second path is task-driven, focusing on designing sophisticated inductive biases for specific tasks, such as the geometric constraints in DSINE~\cite{dsine} for normal estimation.
While efficient, these designs are often difficult to transfer to other perception tasks.
Both paths highlight the need for a powerful and general-purpose visual prior, which is precisely what large-scale generative models can offer.

\subsection{Diffusion Models for Dense Perception}
To leverage the powerful priors of generative models, an emerging line of research adapts diffusion models for dense perception tasks.
Pioneering works like Marigold~\cite{marigold,marigoldv1.1} and GeoWizard~\cite{geowizard} demonstrated the feasibility of fine-tuning pre-trained T2I models (e.g., Stable Diffusion) for depth and normal estimation, achieving excellent generalization with limited data.
This paradigm was subsequently refined by a series of studies: GenPercept~\cite{genpercept}, Stable Normal~\cite{stablenormal}, Lotus~\cite{lotus}, and E2E-FT~\cite{e2e-ft} improved the sampling process or training objectives to enable efficient single-step inference, significantly enhancing the models' practicality.
Similarly, in the domain of image matting, methods such as MAM~\cite{MAM}, SMat~\cite{smat}, and SDMatte~\cite{sdmatte} have also successfully utilized diffusion priors to improve the modeling of fine-grained edge details.

While these works have collectively advanced the field, they are mostly built upon a common foundation: using T2I generative models as the starting point.
In contrast, we explore an alternative possibility.
Based on the ``Representation Mismatch" argument from our introduction, we posit that I2I editing models, pre-trained for structured semantic understanding, can provide a more natural foundation for dense perception tasks.
Concurrent to our work, FE2E~\cite{fe2e} also builds on editing diffusion models, focusing on deterministic zero-noise denoising and consistent-velocity training, with logarithmic depth normalization to address precision issues.
In contrast, our approach emphasizes geometric fidelity, introducing a pixel-space consistency loss and theoretically deriving a square-root depth mapping, while retaining Gaussian noise to better leverage the editor’s generative priors.

\section{Method}

\subsection{Editor2Perceive: Overall Architecture}
Our framework, \textbf{Editor2Perceive}, is built upon FLUX.1 Kontext~\cite{flux-kontext}, a flow-matching~\cite{flowmatching} model based on the DiT~\cite{dit} architecture. FLUX.1 Kontext unifies image generation and editing by incorporating semantic context from both text and image inputs through a simple sequence concatenation mechanism. It processes these inputs with a combination of shared and task-specific transformer blocks to extract image features, and generates new views that maintain semantic consistency, which is vital for dense perception tasks.

We formalize dense perception as a conditional diffusion editing task. As illustrated in Fig.~\ref{fig:framework}, given an input RGB image $x \in \mathbb{R}^{H \times W \times 3}$ and a text prompt $p$, our goal is to predict the corresponding target image $y \in \mathbb{R}^{H \times W \times 3}$. Our method operates in the latent space of a pre-trained VAE. First, we encode the inputs: the conditioning image $x$ is encoded into a latent representation $c_x$, the target dense map $y$ into a target latent $z_1$, and the text prompt $p$ into a text embedding $c_p$.

Next, we define a straight-line trajectory from a standard Gaussian noise vector $z_0 \sim \mathcal{N}(0, \mathbf{I})$ to the target latent $z_1$ using Rectified Flow~\cite{rectified}:
\begin{equation}
    z_t = (1-t)z_0 + t z_1, \quad t \in [0, 1].
    \label{eq:flow_path}
\end{equation}
The constant velocity vector of this path is $\boldsymbol{v} = z_1 - z_0$. Our DiT model is trained to predict this velocity, denoted as $\boldsymbol{v}_\theta$ conditioned on the intermediate state $z_t$, the timestep $t$, and the conditions $c_x$ and $c_p$

\paragraph{Training Objective.} Our model is trained by optimizing a composite loss function that combines a latent-space flow matching with a pixel-space consistency loss. The primary objective is the flow matching loss, which minimizes the mean squared error between the predicted and true velocities:
\begin{equation}
    \mathcal{L}_{\text{FM}} = \mathbb{E}_{t, z_1, z_0, c} \left\| \boldsymbol{v}_\theta(\text{concat}(z_t, c_x, c_p), t) - \boldsymbol{v} \right\|^2_2.
\end{equation}
While $\mathcal{L}_{\text{FM}}$ effectively guides the generation process in the latent space, it lacks direct supervision on the final pixel-level output, which can lead to geometric inaccuracies. To address this, we introduce an pixel-space consistency loss, $\mathcal{L}_{\text{Cons}}$, to enhance geometric fidelity. Our final training objective is a weighted sum of these two losses:
\begin{equation}
    \mathcal{L} = \mathcal{L}_{\text{FM}} + \lambda \mathcal{L}_{\text{Cons}},
\end{equation}
Here, $\lambda$ is an adaptive weight that balances the two objectives. The detailed formulation of $\mathcal{L}_{\text{Cons}}$ for different dense perception tasks and the curriculum-based strategy for weighting $\lambda$ will be elaborated in Section~\ref{sec:3.2}.

\paragraph{Inference.} During inference, we generate the target latent $\hat{z}_1$ by solving the ordinary differential equation (ODE) that defines the flow. Benefiting from the high efficiency inherent to Flow Matching, we can approximate the solution with a single-step Euler integrator:
\begin{equation}
    \hat{z}_1 = z_0 + \boldsymbol{{v}}_\theta(\text{concat}(z_0, c_x, c_p), t=0).
    \label{eq:inference}
\end{equation}
The final dense map $\hat{y}$ is then obtained by the VAE decoder. Furthermore, to enhance the determinism of the process, both training and inference share a fixed random seed for the initial noise $z_0$. Following Marigold~\cite{marigold}, we also adopt an annealed multi-resolution noise strategy.

\subsection{Enhancing Geometric Fidelity with Consistency Loss}
\label{sec:3.2}
Although $\mathcal{L}_{\text{FM}}$ aligns the flow in the latent space, it provides no direct supervision on the final pixel-level output reconstructed by the VAE decoder. Minor errors in the latent space can be amplified after decoding, leading to blurriness or structural artifacts.

To bridge this gap and enhance geometric fidelity, we introduce \textbf{pixel-space consistency loss}, $\mathcal{L}_{\text{Cons}}$, computed directly between the decoded prediction $\hat{y}$ and the ground truth $y$. This loss is customized for each task's specific properties. For simplicity, all formulations below are averaged over all pixels, denoted by $\mathbb{E}$.

\paragraph{Monocular Depth Estimation.} We employ a Scale-and-Shift Invariant L1 Loss. We first align the prediction $\hat{y}$ to the ground truth $y$ via least-squares fitting to obtain $\hat{y}_{\text{align}} = s\hat{y} + t$, and then compute the L1 error:
\begin{equation}
    \mathcal{L}_{\text{Cons}}^{\text{depth}} = \mathbb{E}\left[| y -  \hat{y}_{\text{align}}|\right].
\end{equation}

\paragraph{Surface Normal Estimation.} To ensure accurate normal orientation while maintaining numerical stability, we use a mean angular error based on \textit{atan2}. While the traditional \textit{arccos} method suffers from gradient explosion when vectors are nearly collinear, our approach robustly computes the angle by using the dot and cross products to find its cosine and sine values, respectively:
\begin{equation}
    \mathcal{L}_{\text{Cons}}^{\text{normal}} = \mathbb{E}\left[ \text{atan2}\left( |y \times \hat{y}|, y \cdot \hat{y} \right) \right].
\end{equation}
This formulation is mathematically equivalent to \textit{arccos} but gradient-stable.

\paragraph{Interactive Matting.} Following standard practice~\cite{sepl1loss}, we compute separate L1 loss for the unknown transition region $\mathcal{U}$ and the known foreground / background regions $\mathcal{K}$, allowing model to capture challenging edge details:
\begin{equation}
    \mathcal{L}_{\text{Cons}}^{\text{matting}} = \mathbb{E}_{i \in \mathcal{U}}[|y_i - \hat{y}_i|] + \mathbb{E}_{i \in \mathcal{K}}[|y_i - \hat{y}_i|].
\end{equation}

\paragraph{Adaptive Loss Weighting.} The consistency loss $\mathcal{L}_{\text{Cons}}$ provides a valuable geometric gradient from the pixel space to guide the latent-space flow matching. We combine the two losses with an adaptive weight $\lambda$:
\begin{equation}
    \mathcal{L} = \mathcal{L}_{\text{FM}} + \lambda \mathcal{L}_{\text{Cons}} ,
\end{equation}
The weight $\lambda$ is critical and is designed to implement a curriculum. It is set to zero during the first epoch to prioritize learning from the diffusion prior, and then linearly increases to balance the two objectives. This is formulated as:
\begin{equation}
    \lambda = \frac{\text{sg}(|\mathcal{L}_{\text{FM}}|)}{\text{sg}(|\mathcal{L}_{\text{Cons}}|) + \epsilon} \cdot \max\left(0, \frac{\text{step}}{N_{\text{step}}} - 1\right),
\end{equation}
where `sg' denotes the stop-gradient operator and $N_{\text{step}}$ is the total number of steps in one epoch, $\epsilon$ is a small number (0.001) to aviod devived by zero. This formulation enbales the model focus on latent space flow matching at initial epoch and then gradually turn into a pixel space consistency loss. 

\subsection{Task-Specific Data Representation}
\label{sec:3.3}
To seamlessly adapt the supervision signal $y$ of various dense perception tasks to the input requirements of the pre-trained editing model (a 3-channel BF16 tensor in the $[-1, 1]$ range), we design task-specific pre-processing pipelines aimed at maximizing information preservation and minimizing quantization error.


\paragraph{Monocular Depth Estimation.}
Depth maps ($y$) are single-channel and exhibit a long-tailed distribution. Direct linear normalization of $y$ under BF16 precision causes significant quantization error in near-field details. To address this, we seek an optimal non-linear mapping, $g(y)$, that minimizes the quantization-induced relative error.

We formalize this by minimizing the integral of the relative error over the depth range $[y_{\min}, y_{\max}]$. Detailed in Appendix, this objective simplifies to minimizing the following integral with respect to the function $g$:
\begin{equation}
\small
    \min_{g} \frac{1}{512(y_{\max}-y_{\min})}\int_{y_{\min}}^{y_{\max}} \frac{g(y)_{\max}-g(y)_{\min}}{y \cdot g'(y)} \mathrm{d}y.
    \label{eq:error_integral}
\end{equation}
By applying the Cauchy-Schwarz inequality, we theoretically prove that this integral is minimized when $g'(y) \propto 1/\sqrt{y}$. This yields the optimal mapping function:
\begin{equation}
    g(y) = \sqrt{y}.
    \label{eq:sqrt_mapping}
\end{equation}
Let $y_{\text{sqrt}} = g(y)$. We then apply a robust percentile-based linear normalization to these mapped values to scale them into the $[-1, 1]$ range for the VAE:
\begin{equation}
    y_{\text{norm}} = \left( \frac{y_{\text{sqrt}} - y_{\text{sqrt, p2}}}{y_{\text{sqrt, p98}} - y_{\text{sqrt, p2}}} - 0.5 \right) \times 2.
    \label{eq:robust_norm}
\end{equation}
Here, $y_{\text{sqrt, p2}}$ and $y_{\text{sqrt, p98}}$ represent the 2nd and 98th percentiles of the sqrt-transformed image $y\_\text{sqrt}$. Finally, the single-channel $y_{\text{norm}}$ is replicated to three channels to form the final input representation for our framework.

\paragraph{Surface Normal Estimation.} Normal maps are inherently 3-channel vectors. We simply ensure they are normalized to unit length: $y_{\text{norm}} = y / \|y\|_2$.

\paragraph{Interactive Matting.} Alpha mattes are single-channel images in the $[0, 1]$ range. We first binarize the matte and then linearly map it to the $[-1, 1]$ range:
\begin{equation}
    y_{\text{norm}} = (\mathbb{I}(y > 0.5) - 0.5) \times 2,
\end{equation}
then replicating it to three channels.

\begin{table*}[!h]
    \centering
      \footnotesize
    \setlength{\tabcolsep}{3.7pt}
    \caption{Quantitative comparison for zero-shot monocular depth estimation benchmarks. Our method is evaluated against recent SOTA methods across five benchmarks. The \colorbox{first}{best} and \colorbox{second}{second-best} are highlighted. All metrics are prresented in percentage terms.}
    \label{tab:depth_comp}
    \begin{tabular}{lc *{10}{C{0.97cm}} C{1.4cm}} 
    \toprule
        \multirow{2}{*}{\textbf{Method}} & \multirow{2}{*}{\textbf{Data}} & 
        \multicolumn{2}{c}{\textbf{NYU}} & \multicolumn{2}{c}{\textbf{KITTI}} & 
        \multicolumn{2}{c}{\textbf{ETH3D}} & \multicolumn{2}{c}{\textbf{Scannet}} & 
        \multicolumn{2}{c}{\textbf{DIODE}} & \multirow{2}{*}{\textbf{AvgRank$\downarrow$}} \\
        & & \textbf{AbsRel$\downarrow$} & \textbf{$\delta_1$$\uparrow$} & 
        \textbf{AbsRel$\downarrow$} & \textbf{$\delta_1$$\uparrow$} & 
        \textbf{AbsRel$\downarrow$} & \textbf{$\delta_1$$\uparrow$} & 
        \textbf{AbsRel$\downarrow$} & \textbf{$\delta_1$$\uparrow$} & 
        \textbf{AbsRel$\downarrow$} & \textbf{$\delta_1$$\uparrow$} & \\
        \midrule
        MiDaS & 2M & 11.1 & 88.5 & 23.6 & 63.0 & 18.4 & 75.2 & 12.1 & 84.6 & 33.2 & 71.5 & 9.5 \\
        GeoWizard & 280k & 5.2 & 96.6 & 9.7 & 92.1 & 6.4 & 96.1 & 6.1 & 95.3 & 29.7 & \cellcolor{second} 79.2 & 4.8 \\
        GenPercept & 74k & 5.6 & 96.0 & 9.9 & 90.4 & 6.2 & 95.8 & - & - & 35.7 & 75.6 & 6.0 \\
        DepthAnything V2 & 62.6M & \cellcolor{second} 4.5 & \cellcolor{first} 97.9 & \cellcolor{first} 7.4 & \cellcolor{first} 94.6 & 13.1 & 86.5 & - & - & 26.5 & 73.4 & 3.5 \\
        Marigold & 74k & 5.5 & 96.4 & 9.9 & 91.6 & 6.5 & 95.9 & 6.4 & 95.2 & 30.8 & 77.3 & 5.3 \\
        Marigold1.1 & 74k & 5.5 & 96.4 & 10.5 & 90.2 & 6.9 & 95.7 & 5.8 & 96.3 & 29.8 & 78.2 & 6.4 \\
        Lotus-D & 59k & 5.1 & 97.2 & 8.1 & 93.1 & 6.1 & \cellcolor{second} 97.0 & \cellcolor{second} 5.5 & \cellcolor{second} 96.5 & \cellcolor{first} 22.8 & 73.8 & \cellcolor{second} 2.9 \\
        Lotus-G & 59k & 5.4 & 96.8 & 8.5 & 92.2 & \cellcolor{second} 5.9 & \cellcolor{second} 97.0 & 5.9 & 95.7 & \cellcolor{second} 22.9 & 72.9 & 4.2 \\
        E2E-FT & 74k & 5.4 & 96.5 & 9.6 & 92.1 & 6.4 & 95.9 & 5.8 & \cellcolor{second} 96.5 & 30.3 & 77.6 & 4.7 \\
        \textbf{Edit2Percieve} & 74k & \cellcolor{first} 4.4 & \cellcolor{second} 97.6 & \cellcolor{second} 7.9 & \cellcolor{second} 94.5 & \cellcolor{first} 4.3 & \cellcolor{first} 98.3 & \cellcolor{first} 4.9 & \cellcolor{first} 97.3 & 24.8 & \cellcolor{first} 81.4 & \cellcolor{first} 1.5 \\
    \bottomrule
    \end{tabular}
\end{table*}
\begin{table*}[!h]
    \centering
    \footnotesize
    \caption{Quantitative evaluation of our method on zero-shot surface normal estimation benchmarks. The \colorbox{first}{best} and \colorbox{second}{second-best} performances are highlighted.}
    \label{tab:normal_comp}
    \setlength{\tabcolsep}{9pt}
    \begin{tabular}{l *{10}{c}} 
        \toprule
        \multirow{2}{*}{\textbf{Method}} & \multirow{2}{*}{\textbf{Data}} & 
        \multicolumn{2}{c}{\textbf{NYU}} & \multicolumn{2}{c}{\textbf{Scannet}} & 
        \multicolumn{2}{c}{\textbf{iBims-1}} & \multicolumn{2}{c}{\textbf{DIODE}} & 
        \multirow{2}{*}{\textbf{AvgRank$\downarrow$}} \\
        & & \textbf{Mean$\downarrow$} & \textbf{11.25°$\uparrow$} & \textbf{Mean$\downarrow$} & \textbf{11.25°$\uparrow$} & 
        \textbf{Mean$\downarrow$} & \textbf{11.25°$\uparrow$} & \textbf{Mean$\downarrow$} & \textbf{11.25°$\uparrow$} & \\
        \midrule
        GeoWizard & 278k & 19.0 & 50.0 & 17.6 & 54.6 & 19.3 & 62.3 & 24.7 & 30.1 & 8.3 \\
        GenPercept & 44k & 18.3 & 56.0 & 18.2 & 57.4 & 18.3 & 63.8 & 22.3 & 38.1 & 7.4 \\
        StableNormal & 278k & 17.8 & 54.2 & 16.7 & 54.0 & 17.1 & 67.8 & 19.3 & \cellcolor{first}53.8 & 5.5 \\
        DSINE & 160K & 16.2 & \cellcolor{second}61.0 & 16.4 & 59.6 & 17.1 & 67.4 & 19.9 & 41.8 & 4.5 \\
        Lotus-D & 59k & 16.2 & 59.8 & 14.7 & 64.0 & 17.1 & 66.4 & - & - & 3.1 \\
        Lotus-G & 59k & 16.5 & 59.4 & 15.1 & 63.9 & 17.2 & 66.2 & - & - & 4.4 \\
        ETE-FT & 74k & 16.5 & 60.4 & 14.7 & \cellcolor{second}66.1 & \cellcolor{second}16.1 & \cellcolor{second}69.7 & 19 & 44.4 & 3.0 \\
        Marigold v1.1 & 77k & \cellcolor{second}16.1 & 60.5 & \cellcolor{second}14.5 & \cellcolor{second}66.1 & 16.3 & 68.5 & \cellcolor{second}18.8 & \cellcolor{second}45.5 & \cellcolor{second}2.4 \\
        \textbf{Edit2Percieve} & 77k & \cellcolor{first}15.7 & \cellcolor{first}61.6 & \cellcolor{first}14.1 & \cellcolor{first}66.3 & \cellcolor{first}15.1 & \cellcolor{first}70.9 & \cellcolor{first}18.7 & 44.3 & \cellcolor{first}1.4 \\
        \bottomrule
        \end{tabular}
\end{table*}
\begin{table*}[!h]
\centering
  \footnotesize
\setlength{\tabcolsep}{2.4pt}
\caption{Quantitative evaluation of our method on interactive matting benchmarks. The \colorbox{first}{best} and \colorbox{second}{second-best} performances are highlighted. All the methods are giving same visual input points.}
\label{tab:matting_comp_inter}
\begin{tabular}{l *{5}{c} *{5}{c} *{5}{c} c} 
        \toprule
        \multirow{2}{*}{\textbf{Method}} & \multicolumn{5}{c}{\textbf{AIM-500}} & \multicolumn{5}{c}{\textbf{P3M-500-NP}} & \multicolumn{5}{c}{\textbf{AM-2k}} &  \multirow{2}{*}{\textbf{AvgRank$\downarrow$}} \\
        \cmidrule(lr){2-6} \cmidrule(lr){7-11} \cmidrule(lr){12-16}  
        & \textbf{MSE$\downarrow$} & \textbf{MAD$\downarrow$} & \textbf{SAD$\downarrow$}  & \textbf{Grad$\downarrow$} & \textbf{Conn$\downarrow$} & \textbf{MSE$\downarrow$} & \textbf{MAD$\downarrow$} & \textbf{SAD$\downarrow$} & \textbf{Grad$\downarrow$} & \textbf{Conn$\downarrow$} & \textbf{MSE$\downarrow$} & \textbf{MAD$\downarrow$} & \textbf{SAD$\downarrow$} & \textbf{Grad$\downarrow$} & \textbf{Conn$\downarrow$} & \\
        
        \midrule
        MAM & 0.075 & 0.108 & 186.5 & 37.5 & 40.4 & 0.087 & 0.116 & 207.5 & 29.4 & 43.5 & 0.060 & 0.081 & 141.6 & 22.5 & 31.5 & 5.9 \\
        MatAny & 0.043 & 0.052 & 87.0 & 33.4 & 25.4 & 0.029 & 0.034 & 57.3 & 25.9 & \cellcolor{second}16.0 & 0.012 & 0.019 & 32.2 & 15.7 & 20.4 & 4.4 \\
        SmartMatting & 0.030 & 0.039 & 66.3 & 46.6 & 18.8 & 0.024 & 0.029 & 50.5 & 28.5 & 19.6 & 0.030 & 0.037 & 62.6 & 33.8 & 15.9 & 4.3 \\
        LiteSDMatte & 0.011 & 0.021 & 34.4 & \cellcolor{second}24.3 & 20.0 & \cellcolor{second}0.012 & \cellcolor{second}0.017 & \cellcolor{second}29.9 & \cellcolor{second}16.6 & 21.8 & 0.009 & 0.016 & 27.5 & 13.6 & 17.7 & 2.9 \\
        SDMatte & \cellcolor{second}0.011 & \cellcolor{second}0.019 & \cellcolor{second}31.8 & 26.8 & \cellcolor{second}17.5 & 0.013 & 0.018 & 32.0 & 20.4 & 20.8 & \cellcolor{second}0.006 & \cellcolor{first}0.010 & \cellcolor{first}17.5 & \cellcolor{second}13.2 & \cellcolor{second}10.9 & \cellcolor{second}2.2 \\
        \textbf{Edit2Percieve} & \cellcolor{first}0.006 & \cellcolor{first}0.017 & \cellcolor{first}29.1 & \cellcolor{first}18.2 & \cellcolor{first}15.7 & \cellcolor{first}0.003 & \cellcolor{first}0.011 & \cellcolor{first}19.4 & \cellcolor{first}13.2 & \cellcolor{first}10.2 & \cellcolor{first}0.004 & \cellcolor{second}0.012 & \cellcolor{second}20.4 & \cellcolor{first}9.6 & \cellcolor{first}9.9 & \cellcolor{first}1.2 \\
        \bottomrule
\end{tabular}

\end{table*}

\section{Experiment}

\subsection{Experiment Setup}

\paragraph{Implementation Details.}
Our experimental framework is built upon the FLUX.1 Kontext backbone. 
All parameters are frozen except DiT.  For depth estimation and normal estimation tasks, we adopt the AdamW optimizer with a learning rate of $3 \times 10^{-5}$ and a batch size of 16. For interactive matting, due to the higher memory demand from paired context images, we use the memory-efficient AdamW8bit optimizer with a batch size of 16.  
To accelerate convergence, we apply annealed multi-resolution noise during training.  
The model typically converges within approximately 6000 training steps.  
Each task is trained on a single NVIDIA H200 GPU for about 1.5 days.



\begin{figure*}
\centering
\begin{subfigure}[t]{0.46\textwidth}
\centering
\includegraphics[height=6cm, width=\linewidth]{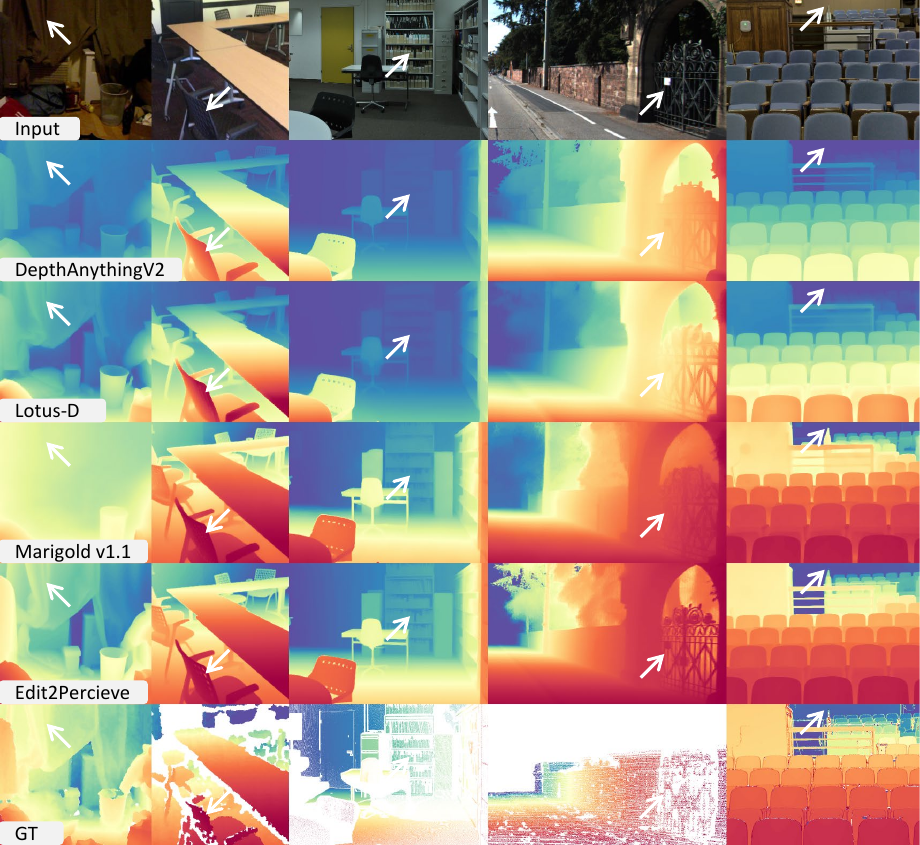}
\caption{Monocular Depth Estimation}
\label{fig:Qualitive of Depth}
\end{subfigure}%
\hfill
\begin{subfigure}[t]{0.27\textwidth}
\centering
\includegraphics[height=6cm, width=\linewidth]{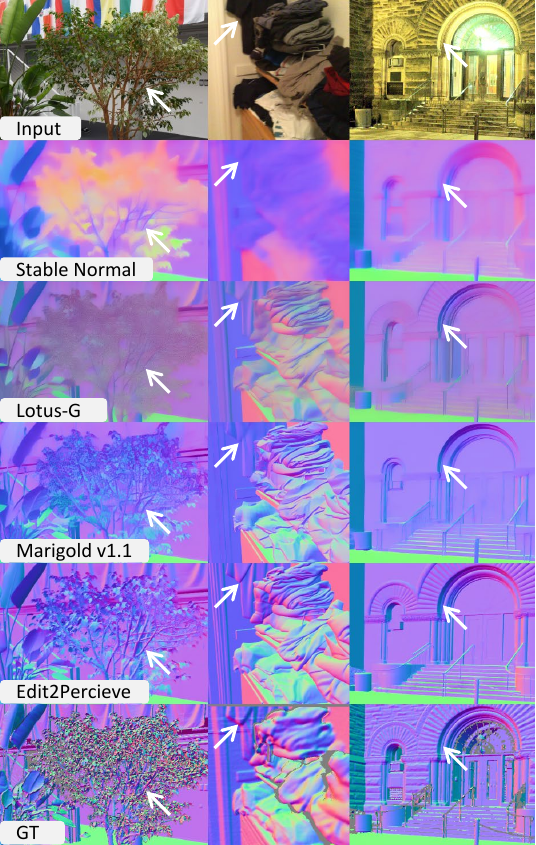}
\caption{Surface Normal Estimation}
\label{fig:Qualitive of Normal}
\end{subfigure}%
\hfill
\begin{subfigure}[t]{0.23\textwidth}
\centering
\includegraphics[height=6cm, width=\linewidth]{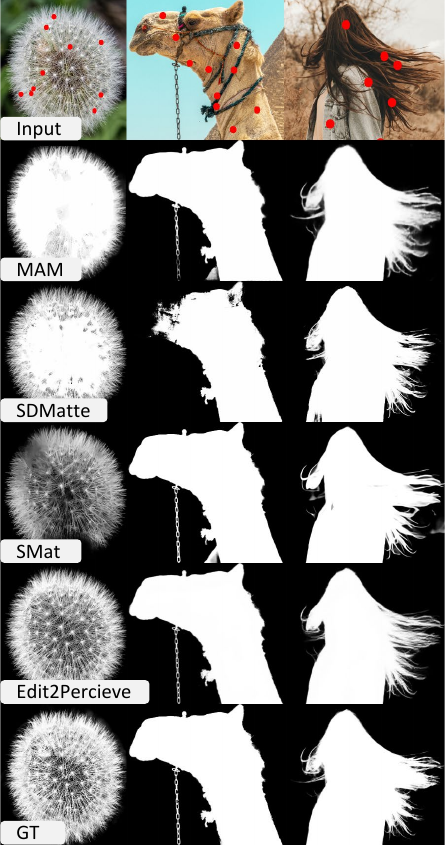}
\caption{Interactive Matting}
\label{fig:Qualitive of Matting}
\end{subfigure}
\caption{Qualitative Comparison of our methods with other SOTA methods across different benchmarks. The arrows emphasize the regions that Edit2Percieve (ours) significantly outperform others. Zoom in for better view.}
\label{fig:Qualitative Evaluation}
\end{figure*}

\paragraph{Dataset and Benchmarks.}

We train and evaluate our model on task-specific benchmark datasets.
\textbf{Monocular Depth Estimation:} Trained on Hypersim~\cite{hypersim} and Virtual KITTI 2~\cite{vkitti2}, with a 90\%:10\% sample ratio.
\textbf{Surface Normal Estimation:} Trained on Hypersim~\cite{hypersim}, InteriorVerse~\cite{iv}, and Sintel~\cite{Sintel}, with a 50\%:45\%:5\% sample ratio.
\textbf{Interactive Matting:} Trained on AM-2k~\cite{am-2k}, Distinctions-646~\cite{distinctions-646}, Composition-1k~\cite{comp-1k}, and COCO-Matting~\cite{cocomatte}, with a 25\%:25\%:25\%:25\% sample ratio.

For evaluation, we test our models on zero-shot generalization on all benchmarks except for AM-2k in Interactive Matting task.
\textbf{Monocular Depth Estimation:} Evaluated on NYUv2~\cite{nyuv2}, KITTI~\cite{kitti}, ETH3D~\cite{eth3d}, ScanNet~\cite{scannet}, and DIODE~\cite{diode}.
\textbf{Surface Normal Estimation:} Evaluated on NYUv2~\cite{nyuv2}, ScanNet~\cite{scannet}, iBims-1~\cite{ibims-1}, and DIODE~\cite{diode}.
\textbf{Interactive Matting:} Evaluated on P3M-500-NP~\cite{p3m-10k}, AM-2k~\cite{am-2k}, and AIM-500~\cite{aim-500}.

\paragraph{Metrics.}
For Monocular Depth Estimation, we report ~\textit{AbsRel} and $\delta_1$.  
Specifically, $\text{AbsRel} = \mathbb{E}[(\hat{y}_{align}-y)/y]$  
where $\mathbb{E}$ is the expectation (average) of all valid pixels, $\hat{y}_{align}$ denotes the aligned prediction, and $y$ the ground-truth depth.  
$\delta_1$ measures the proportion of pixels satisfying $\max(\hat{y}_{align}/y,\quad y/\hat{y}_{align}) < 1.25$.  

For Surface Normal Estimation, we report $\text{Mean}$ and $\text{11.25}$ the \textit{mean angular error} and the percentage of pixels with error $<11.25^{\circ}$.  

For Interactive Matting, we evaluate using MSE, MAD, SAD, Gradient, and Connectivity metrics.

\subsection{Quantitative Evaluation}

Table~\ref{tab:depth_comp},~\ref{tab:normal_comp}, and~\ref{tab:matting_comp_inter} demonstrate the quantitative results across three dense perception tasks. Despite using far less training data and a single-step inference, our method consistently outperforms all strong baselines, demonstrating the effectiveness of \textbf{image-editing diffusion models} in structural understanding and geometric reasoning.

For Zero-Shot Monocular Depth Estimation, as shown in Table~\ref{tab:depth_comp}, our approach achieves new SOTA performance on five benchmarks, reaching an average rank of $1.5$. On ETH3D and Scannet, the AbsRel error decreases $27\%$ and $11\%$ compared to second-best model, respectively. Notably, our model is trained on only $74\text{K}$ images yet surpasses heavily supervised approaches such as \textit{DepthAnything V2} ($62.6\text{M}$ images), highlighting the efficiency of our depth representation optimization.

For Zero-Shot Surface Normal Estimation (Table~\ref{tab:normal_comp}), our model ranks first on all benchmarks with an average rank of $1.4$, confirming strong cross-dataset robustness and superior geometric detail reconstruction.

Finally, as shown in Table~\ref{tab:matting_comp_inter}, our interactive variant achieves the lowest errors on all three datasets and an average rank of $1.2$, demonstrating precise alpha prediction and excellent generalization to both portrait and non-portrait scenes.

\subsection{Qualitative Evaluation}

In this section, we present the qualitative evaluation results.  As shown in Fig.~\ref{fig:Qualitative Evaluation}, we compare our method with baseline approaches on the Monocular Depth Estimation, Surface Normal Esitmation and Interactive Matting. 
Our model produces depth maps with finer structural details and stronger consistency with the input image, while baseline methods tend to miss or blur small but critical objects.  

Similar improvements are observed in the surface normal and matting tasks.  
Our results exhibit better spatial coherence and visual consistency, demonstrating the model’s capability to preserve both geometric accuracy and perceptual quality across dense perception tasks.

\definecolor{purple-custom}{RGB}{128,0,128} 
\definecolor{yellow-custom}{RGB}{204,204,0} 
\definecolor{green-custom}{RGB}{0,128,0}    

\subsection{Ablation Study}
\label{sec:ablation}

To thoroughly analyze the contributions of each component in our framework, we conduct a series of detailed ablation studies. We first validate our core thesis on the superiority of I2I editing models as a foundation, then separately evaluate the effectiveness of our proposed pixel-space consistency loss and the theoretically optimal depth mapping function.

\paragraph{Importance of the Base Model: Image-to-Image (I2I) vs. Text-to-Image (T2I).}
Our core thesis posits that due to differences in pre-training objectives, I2I models designed for editing are better suited for dense perception than T2I generators. To provide direct experimental evidence, we conduct a rigorous controlled experiment.

\textit{Experiment Setup.} We select two models with identical architectures: \textbf{FLUX.1 Kontext (I2I)} and \textbf{FLUX.1 (T2I)}. Crucially, we construct the same I2I-style fine-tuning pipeline for the T2I model (FLUX.1) as our main framework, using token concatenation for the text prompt, condition image and target image. This ensures that the only significant difference between the two models is the prior knowledge acquired during their respective pre-training phases.

\textit{Results and Analysis.} As shown in Table~\ref{tab:ablation_depth} and Table~\ref{tab:ablation_normal}, we conducted a total of six sets of controlled experiments (four for depth estimation and two for normal estimation). The results are unequivocal: across all experimental settings, the performance of the I2I-based model comprehensively and significantly surpasses its T2I counterpart.
For instance, in the depth estimation task (Table~\ref{tab:ablation_depth}), even with the most basic configuration (ID 5$\And$1), the I2I model achieves a remarkable \textbf{25\%} and \textbf{27\%} relative improvement in AbsRel on NYUv2 and KITTI, respectively. This advantage persists even after applying all our optimization strategies (ID 8$\And$4). The same trend is observed in the normal estimation task (Table~\ref{tab:ablation_normal}). This overwhelming experimental evidence strongly validates our core thesis: the structured semantic priors learned by I2I models provide a far superior starting point for downstream perception tasks than those from T2I models.

\paragraph{Effect of Pixel-Space Consistency Loss.}
In Section~\ref{sec:3.2}, we proposed the pixel-space consistency loss ($\mathcal{L}_{\text{Cons}}$) to bridge the gap between latent-space supervision and pixel-level quality. To verify its general effectiveness, we conducted six sets of controlled experiments across three tasks and two base models.

\begin{figure}[htbp]
\centering
\footnotesize
\includegraphics[width=\linewidth]{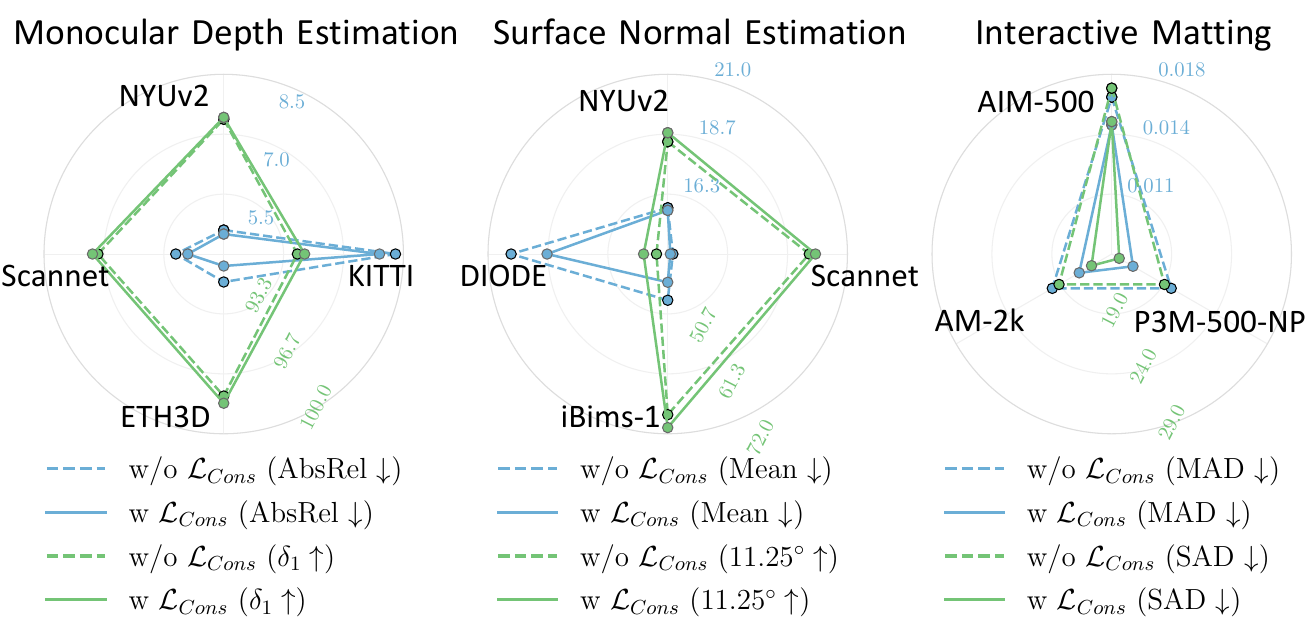}
\caption{\textbf{Effectiveness of the Pixel-Space Consistency Loss ($\mathcal{L}_{\text{Cons}}$) across All Tasks.} The radar charts compare the performance with (solid line) and without (dashed line) our consistency loss. For each task, axes represent key metrics on different datasets (lower is better for error metrics like AbsRel, Mean, MAD, SAD; higher is better for accuracy metrics like $\delta_1$, 11.25$^{\circ}$).}
\label{fig:Ablation_Cons}
\end{figure}

As shown in Fig.~\ref{fig:Ablation_Cons} and Table~\ref{tab:ablation_depth},\ref{tab:ablation_normal}, introducing $\mathcal{L}_{\text{Cons}}$ brings consistent performance improvements across all configurations. Interestingly, we observe a complementary relationship between the gains from $\mathcal{L}_{\text{Cons}}$ and the strength of the base model. For example, on the weaker T2I backbone (Table~\ref{tab:ablation_depth}, IDs 1$\And$2, 3$\And$4), $\mathcal{L}_{\text{Cons}}$ provides substantial improvements (AbsRel drop of 1.0-1.4 on NYUv2). On the stronger I2I backbone (IDs 5$\And$6, 7$\And$8), where the model already possesses better structural understanding, the gains become more subtle (AbsRel drop of 0.3-0.4), indicating that $\mathcal{L}_{\text{Cons}}$ acts more as a \textbf{fine-tuner} rather than a \textbf{rectifier}.
Furthermore, as illustrated in Fig.~\ref{fig:Ablation_Cons}, this improvement is crucial for tasks highly sensitive to fine-grained structures and edges, such as normal estimation and image matting. This proves that $\mathcal{L}_{\text{Cons}}$ serves as an effective plug-and-play module that injects pixel-level geometric constraints into the latent-space generation process, thereby enhancing the final perception quality.

\paragraph{Effect of Theoretically Optimal Depth Mapping.}
In Section~\ref{sec:3.3}, we theoretically derived the optimality of the square root (Sqrt) depth mapping for minimizing relative quantization error.

We conducted four sets of controlled experiments in Table~\ref{tab:ablation_depth} (IDs 1$\And$3, 2$\And$4, 5$\And$7, 6$\And$8) to compare our Sqrt mapping against traditional uniform normalization (Uni). The results show that the Sqrt mapping yields significant performance improvements in all cases.

We observe that the improvement is particularly pronounced on the outdoor dataset KITTI, which features a larger depth range. For instance, without $\mathcal{L}_{\text{Cons}}$ (IDs 1$\And$3, 5$\And$7), the AbsRel reduction on KITTI (-3.0, -1.4) is much larger than on the indoor dataset NYUv2 (-0.5, -0.4). This phenomenon aligns perfectly with our theoretical analysis. As shown in formulation~\ref{eq:error_integral}, our integral error analysis, based on the objective
predicts this behavior. By substituting $g(y)=y$ (Uniform) and $g(y)=\sqrt{y}$ (Sqrt) into the integral for different depth ranges, we can quantify the theoretical error reduction. For an indoor range like NYUv2 ($[0.1, 10]$m), the theoretical improvement for AbsRel is 0.26. For an outdoor range like KITTI ($[0.1, 80]$m), the theoretical improvement for AbsRel is a more substantial 0.6. The strong consistency between our experimental results and theoretical predictions not only proves the superiority of our depth mapping method but also highlights that a principled, first-principles-based design is crucial for unlocking the full potential of large-scale models and achieving SOTA performance.

\begin{figure}
    \centering
    \includegraphics[width=\linewidth]{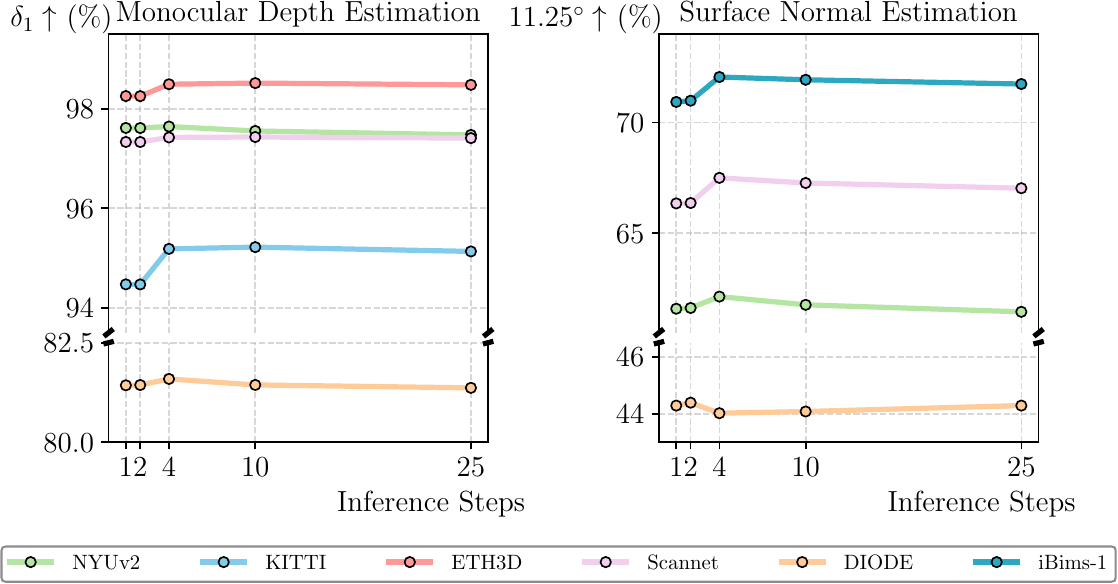}
    \caption{Ablation study on inference steps.}
    \label{fig:Ablation_infer_steps}
\end{figure}

\paragraph{Effect of Inference Steps.} Our framework inherently supports efficient single-step inference. As shown in Fig.~\ref{fig:Ablation_infer_steps}, performance peaks at just 4 steps and then slightly degrades. This suggests that unlike generative tasks, excessive steps can introduce over-smoothing artifacts, reinforcing that our deterministic, single-step approach is optimally aligned with dense perception.
\begin{table}[htbp]
  \centering
  \footnotesize
  \setlength{\tabcolsep}{2.5pt} 
  \caption{Ablation on the Base Model, Consistency Loss, and Depth Normalization for Monocular Depth Estimation. Here the column ``D.M." stands for Depth Mapping, we compare two ways: Uni (Uniform) and Sqrt (Square Root). The \colorbox{first}{best} and \colorbox{second}{second-best} performances are highlighted.}
  \label{tab:ablation_depth}
    \begin{tabular}{llcccccc}
    \toprule
    \multirow{2}{*}{\textbf{ID}} & \multirow{2}{*}{\textbf{Base Model}} & \multirow{2}{*}{\textbf{$\mathcal{L}_{Cons}$}} & \multirow{2}{*}{\textbf{D.M.}} & 
    \multicolumn{2}{c}{\textbf{NYUv2}} & \multicolumn{2}{c}{\textbf{KITTI}} \\
    & & & & \textbf{AbsRel $\downarrow$} & \textbf{$\delta_1$ $\uparrow$} & \textbf{AbsRel $\downarrow$} & \textbf{$\delta_1$ $\uparrow$} \\
    \midrule
    1 & FLUX.1 &  & Uni & 6.8 & 95.1 & 13.2 & 83.7 \\
    2 & FLUX.1 & \checkmark & Uni & 5.4 & 96.9 & 12.5 & 84.3 \\
    3 & FLUX.1 &  & Sqrt & 6.3 & 95.8 & 10.2 & 89.7 \\
    4 & FLUX.1 & \checkmark & Sqrt & 5.3 & 97.0 & 8.4 & 92.8 \\
    5 & FLUX.1 Kontext &  & Uni & 5.1 & 96.9 & 9.6 & 91.2 \\
    6 & FLUX.1 Kontext & \checkmark & Uni & 4.8 & 97.2 & 9.6 & 91.2 \\
    7 & FLUX.1 Kontext &  & Sqrt & \cellcolor{second}4.7 & \cellcolor{second}97.5 & \cellcolor{second}8.2 & \cellcolor{second}94.1 \\
    8 & FLUX.1 Kontext & \checkmark & Sqrt & \cellcolor{first}4.4 & \cellcolor{first}97.6 & \cellcolor{first}7.9 & \cellcolor{first}94.5 \\
    \bottomrule
    \end{tabular}%
\end{table}%

\begin{table}[htbp]
  \centering
  \footnotesize
  \setlength{\tabcolsep}{3pt} 
  \caption{Ablation on the Base Model, Consistency Loss for Surface Normal Estimation. The \colorbox{first}{best} and \colorbox{second}{second-best} performances are highlighted.}
  \label{tab:ablation_normal} 
    \begin{tabular}{ccccccc}
    \toprule
    \multirow{2}{*}{\textbf{ID}} & \multirow{2}{*}{\textbf{Base Model}} & \multirow{2}{*}{\textbf{$\mathcal{L}_{Cons}$}} & \multicolumn{2}{c}{\textbf{NYUv2}} & \multicolumn{2}{c}{\textbf{Scannet}} \\
    & & & \textbf{Mean $\downarrow$} & \textbf{11.25° $\uparrow$} & \textbf{Mean $\downarrow$} & \textbf{11.25° $\uparrow$} \\
    \midrule
    1 & FLUX.1 &  & 16.6 & 57.7 & 15.0 & 62.1 \\
    2 & FLUX.1 & \checkmark & 16.4 & 59.1 & 14.9 & 63.3 \\
    3 & FLUX.1 Kontext &  & \cellcolor{second}15.8 & \cellcolor{second}60.0 & \cellcolor{second}14.2 & \cellcolor{second}65.2 \\
    4 & FLUX.1 Kontext & \checkmark & \cellcolor{first}15.7 & \cellcolor{first}61.6 & \cellcolor{first}14.1 & \cellcolor{first}66.3 \\
    \bottomrule
    \end{tabular}%
\end{table}%

\subsection{Future Work}
Our approach shows strong potential for dense perception tasks, but future work could extend to sparse tasks, such as human pose estimation and object detection. Furthermore, improving computational efficiency will be essential, as our current method demands considerable resources due to the computationally intensive DiT blocks.

\section{Conclusion}

In this work, we presented \textbf{Editor2Perceive}, a simple yet powerful diffusion-transformer framework for dense perception tasks.  
Unlike traditional text-to-image diffusion models, our approach leverages image-to-image diffusion models as geometric priors, reframing dense perceptioin as a deterministic image-to-image transformation process.  
This perspective allows the model to achieve strong spatial consistency and structural fidelity across diverse tasks such as Monocular Depth Estimation, Surface Normal Estimation, and Interactive Matting. By introducing pixel-space consistency loss and efficient single-step inference, Editor2Perceive not only achieves SOTA accuracy but also significantly reduces inference cost.  
Our results demonstrate that diffusion-based editors can serve as a new class of perception-oriented foundation models, combining the expressive power of generative models with the precision and stability required for geometric reasoning.  
We believe this work opens up promising directions for unifying editing and perception within a single diffusion framework, paving the way for efficient, structure-aware, and general-purpose visual understanding.

{
    \small
    \bibliographystyle{unsrtnat}
    \bibliography{main}
}

\appendix

\clearpage
\setcounter{page}{1}
\maketitlesupplementary

\section{Training Details}
\subsection{Preprocess of input RGB image}
\label{appendix:rgb_preprocess}

For all tasks, the input RGB image $x$, typically in 3-channel uint8 format, is linearly normalized to the range $[-1, 1]$ to match the VAE's input requirements.

For the Interactive Matting task, we incorporate an additional visual prompt in the form of user-provided points. Following~\cite{sdmatte}, we simulate these points during training by randomly sampling up to 10 points from the foreground region. A soft mask $M_p \in [0, 1]^{H \times W}$ is then generated by placing a Gaussian kernel at each point's coordinates. This mask is also normalized to $[-1, 1]$, encoded by the VAE, and its latent representation is concatenated with the other input tokens, serving as an extra condition for the model.

\subsection{Derivation of the Optimal Depth Mapping Function}
\label{sec:appendix_derivation}

This appendix provides a detailed derivation for the optimal non-linear mapping $g(y)$ that minimizes the quantization-induced relative error, as presented in Equation~\ref{eq:error_integral} of the main text.

\paragraph{Problem Formulation.}
The process of converting physical depth values into a model-compatible format involves four distinct stages:
$y \xrightarrow{\text{mapping } g} z \xrightarrow{\text{normalization}} d \xrightarrow{\text{quantization}} q$.

Here, $y$ represents the physical depth, $z=g(y)$ is the depth value after applying the non-linear mapping $g$, $d \in [-1, 1]$ is the value after normalization, and $q$ is the final representation quantized to BF16 precision. Our objective is to determine the mapping function $g(y)$ that minimizes the relative error $\Delta y / y$ propagated back from the final quantization step.

\paragraph{Derivation.}
The BF16 (bfloat16) floating-point format uses 1 sign bit, 8 exponent bits, and 7 fraction (mantissa) bits. For a normalized value $d \in [-1, 1]$, the exponent is at most 127 (representing values up to $2^0=1$). The largest quantization step, $\Delta d$, for values in the range $(-1, 1)$ occurs when the exponent is 126 (for values in $[0.5, 1)$), resulting in:
\begin{equation}
\small
    \Delta d = 2^{(\text{exponent}-127)} \cdot 2^{-7} = 2^{(126-127)} \cdot 2^{-7} = 2^{-8} = \frac{1}{256}.
\end{equation}
This quantization error $\Delta d$ propagates backward through the preceding stages. The linear normalization maps the range of the function $g$, denoted as $[z_{\min}, z_{\max}] = [g(y_{\min}), g(y_{\max})]$, to the interval $[-1, 1]$. The normalization is defined as $d = \frac{z - z_{\min}}{z_{\max} - z_{\min}} \cdot 2 - 1$. The error in the mapped space, $\Delta z$, is therefore:
\begin{equation}
    \Delta z = \frac{z_{\max} - z_{\min}}{2} \Delta d = \frac{g(y_{\max}) - g(y_{\min})}{512}.
\end{equation}
Using the chain rule, we can express the error in the original physical depth space, $\Delta y$, as $\Delta y \approx \frac{dy}{dz} \Delta z$. Since $z=g(y)$, we have $\frac{dz}{dy} = g'(y)$, which implies $\frac{dy}{dz} = \frac{1}{g'(y)}$. Our goal is to minimize the relative error, $\Delta y / y$, across the entire depth range $[y_{\min}, y_{\max}]$. The error at any given point $y$ is:
\begin{equation}
    \frac{\Delta y}{y} = \frac{1}{y} \frac{dy}{dz} \Delta z = \frac{1}{y \cdot g'(y)} \frac{g(y_{\max}) - g(y_{\min})}{512}.
\end{equation}
To find the optimal function $g$ that minimizes this error over the continuous range, we formulate the problem as the minimization of the average relative error, which is equivalent to minimizing its integral:
\begin{equation}
    \min_{g} \int_{y_{\min}}^{y_{\max}} \frac{1}{y \cdot g'(y)} \mathrm{d}y \cdot \left[ g(y_{\max}) - g(y_{\min}) \right].
    \label{eq:appendix_objective}
\end{equation}
(Note: This expression, up to a constant factor, is what is presented in Equation~\ref{eq:error_integral}).

We can rewrite the term $g(y_{\max}) - g(y_{\min})$ as the integral of its derivative: $\int_{y_{\min}}^{y_{\max}} g'(y) \mathrm{d}y$. Substituting this into our objective function gives:
\begin{equation}
    \min_{g} \left( \int_{y_{\min}}^{y_{\max}} g'(y) \mathrm{d}y \right) \left( \int_{y_{\min}}^{y_{\max}} \frac{1}{y \cdot g'(y)} \mathrm{d}y \right).
\end{equation}
This expression is in the form of the product of two integrals, which can be addressed using the Cauchy-Schwarz inequality for integrals. The inequality states that for any two functions $A(y)$ and $B(y)$:
$(\int A(y)B(y) \mathrm{d}y)^2 \le (\int A(y)^2 \mathrm{d}y)(\int B(y)^2 \mathrm{d}y)$.

Let's define $A(y) = \sqrt{g'(y)}$ and $B(y) = \frac{1}{\sqrt{y \cdot g'(y)}}$. Then:
\begin{itemize}
    \item $\int A(y)^2 \mathrm{d}y = \int g'(y) \mathrm{d}y$
    \item $\int B(y)^2 \mathrm{d}y = \int \frac{1}{y \cdot g'(y)} \mathrm{d}y$
\end{itemize}
The product of these two integrals is minimized when the equality in the Cauchy-Schwarz inequality holds. This occurs if and only if one function is a constant multiple of the other, i.e., $A(y) = k \cdot B(y)$ for some constant $k$.
\begin{align}
    \sqrt{g'(y)} &= k \cdot \frac{1}{\sqrt{y \cdot g'(y)}} \\
    \implies g'(y) &= \frac{k^2}{y \cdot g'(y)} \\
    \implies (g'(y))^2 &= \frac{k^2}{y} \\
    \implies g'(y) &\propto \frac{1}{\sqrt{y}}.
\end{align}
Integrating this result with respect to $y$ yields the optimal form for the mapping function $g(y)$:
\begin{equation}
    g(y) \propto \sqrt{y}.
\end{equation}
This derivation proves that a square-root mapping is theoretically optimal for minimizing the relative quantization error when representing depth values under BF16 precision.

\subsection{Derivation of the Numerically Stable Normal Consistency Loss}
\label{appendix:normal_loss}

Our pixel-space consistency loss for surface normal estimation, as presented in Section~\ref{sec:3.2}, is designed for numerical stability during training. A naive approach to compute the mean angular error between the ground-truth normal $y$ and the predicted normal $\hat{y}$ (assuming both are unit vectors) is to use the arccosine function:
\begin{equation}
    \mathcal{L}_{\text{naive}} = \mathbb{E}\left[ \arccos(y \cdot \hat{y}) \right].
\end{equation}
However, the derivative of the \textit{arccos} function is given by:
\begin{equation}
    \frac{d}{dx} \arccos(x) = -\frac{1}{\sqrt{1 - x^2}}.
    \label{eq:arccos_derivative}
\end{equation}
As the argument $x = y \cdot \hat{y}$ approaches $\pm 1$ (i.e., when the predicted normal is very accurate and nearly collinear with the ground truth), the denominator of Eq.~\ref{eq:arccos_derivative} approaches zero. This causes the gradient to explode, leading to numerical instability and training divergence.

To circumvent this issue, we adopt a more robust formulation based on the two-argument arctangent function, \textit{atan2}. The angle $\theta$ between two unit vectors can be uniquely determined by its sine and cosine values, which correspond to the magnitude of their cross product and their dot product, respectively:
\begin{align}
    \sin(\theta) &= \|y \times \hat{y}\|_2, \\
    \cos(\theta) &= y \cdot \hat{y}.
\end{align}
Our final loss, as used in the main paper, is then formulated as:
\begin{equation}
    \mathcal{L}_{\text{Cons}}^{\text{normal}} = \mathbb{E}\left[ \text{atan2}\left( \|y \times \hat{y}\|_2, \quad y \cdot \hat{y} \right) \right].
\end{equation}
The \textit{atan2} function has well-defined, bounded gradients across its entire domain, which resolves the instability issue and significantly improves training stability for the normal estimation task.
\section{Additional Quantitive Results}

\label{appendix:experiments}

This appendix provides the complete quantitative results for the ablation studies discussed in the main paper.

\subsection{Detailed Ablation Studies}
For brevity, the main paper analyzes the impact of the base model, consistency loss, and depth mapping on a subset of datasets. Here, we present the full results across all benchmark datasets.

Tables~\ref{tab:ablation_depth_supp}, \ref{tab:ablation_normal_supp}, and~\ref{tab:ablation_matting_supp} provide the comprehensive ablation results for monocular depth estimation, surface normal estimation, and interactive matting, respectively. These tables serve as a supplement to Tables~\ref{tab:ablation_depth}, \ref{tab:ablation_normal}, and Figure~\ref{fig:Ablation_Cons} in the main text.

The complete results confirm the conclusions drawn in the main paper: 
\begin{enumerate*}[label=(\roman*)]
    \item the I2I-based model (FLUX.1 Kontext) consistently outperforms the T2I-based model (FLUX.1);
    \item the pixel-space consistency loss ($\mathcal{L}_{\text{Cons}}$) brings universal performance improvements;
    \item our theoretically optimal square root depth mapping (Sqrt) is significantly superior to uniform normalization (Uni).
\end{enumerate*}

\subsection{Analysis of Inference Steps}
The complete results for the analysis of inference steps are provided in Tables~\ref{tab:ablation_steps_depth_supp}, \ref{tab:ablation_steps_normal_supp}, and~\ref{tab:ablation_steps_matting_supp}. Our framework demonstrates highly efficient inference capabilities. For all experiments reported in the main paper, we use single-step inference by default. As shown in the tables, the performance degradation from using a single step compared to multiple steps is minor and acceptable, confirming the effectiveness of our efficient approach.

\begin{table*}[!h]
  \centering
  \footnotesize
  \setlength{\tabcolsep}{5.8pt} 
  \caption{Additional Ablation Study on the Base Model, Consistency Loss, and Depth Normalization for Monocular Depth Estimation. Here the column ``D.M." stands for Depth Mapping, we compare two ways: Uni (Uniform) and Sqrt (Square Root). The \colorbox{first}{best} and \colorbox{second}{second-best} performances are highlighted.}
  \label{tab:ablation_depth_supp}
    \begin{tabular}{llcccccccccccc}
    \toprule
    \multirow{2}{*}{\textbf{ID}} & \multirow{2}{*}{\textbf{Base Model}} & \multirow{2}{*}{\textbf{$\mathcal{L}_{Cons}$}} & \multirow{2}{*}{\textbf{D.M.}} & 
    \multicolumn{2}{c}{\textbf{NYUv2}} & \multicolumn{2}{c}{\textbf{KITTI}} & \multicolumn{2}{c}{\textbf{ETH3D}} & \multicolumn{2}{c}{\textbf{Scannet}} & \multicolumn{2}{c}{\textbf{DIODE}}\\
    & & & & \textbf{AbsRel $\downarrow$} & \textbf{$\delta_1$ $\uparrow$} & \textbf{AbsRel $\downarrow$} & \textbf{$\delta_1$ $\uparrow$} & \textbf{AbsRel $\downarrow$} & \textbf{$\delta_1$ $\uparrow$} & \textbf{AbsRel $\downarrow$} & \textbf{$\delta_1$ $\uparrow$} & \textbf{AbsRel $\downarrow$} & \textbf{$\delta_1$ $\uparrow$}  \\
    \midrule
        1 & Flux.1 &  & Uni  & 6.8 & 95.1  & 83.7 & 13.2 & 7.4 & 94.7 & 8.3 & 92.7 & 30.1 & 77.2 \\
        2 & Flux.1 & \checkmark & Uni & 5.4 & 96.9 &  84.3 & 12.5 & 6.3 & 95.4 & 6.1 & 96.2 & 29.3 & 77.5 \\
        3 & Flux.1 &  & Sqrt & 6.3 & 95.8 & 89.7 & 10.2 & 6.3 & 96.6 & 7.5 & 93.8 & 26.4 & 78.9 \\
        4 & Flux.1 & \checkmark & Sqrt & 5.3 & 97.0 & 92.8 & 8.4 & 5.7 & 97.1 & 6.5 & 95.9 & 25.5 & 79.8 \\
        5 & Flux.1 Kontext &  & Uni & 5.1  & 96.9 & 91.2 & 9.6 & 5.4 & 96.5 & 5.2 & 96.8 & 29.2 & 78.9 \\
        6 & Flux.1 Kontext & \checkmark  & Uni & 4.8 & 97.2  & 91.2 & 9.6 & 5.3 & 96.9 & 5.3 & 96.7 & 28.9 & 79.3 \\
        7 & Flux.1 Kontext &  & Sqrt & \cellcolor{second}4.7 & \cellcolor{second}97.5  & \cellcolor{second}94.1 & \cellcolor{second}8.2 & \cellcolor{second}4.7 & \cellcolor{second}98.0 & \cellcolor{second}5.3 & \cellcolor{second}97 & \cellcolor{second}25.2 & \cellcolor{second}81.0 \\
        8 & Flux.1 Kontext & \checkmark  & Sqrt & \cellcolor{first}4.4  &\cellcolor{first}97.6 & \cellcolor{first}94.5 & \cellcolor{first}7.9 & \cellcolor{first}4.3 & \cellcolor{first}98.3 & \cellcolor{first}4.9 & \cellcolor{first}97.3 & \cellcolor{first}24.8 & \cellcolor{first}81.4 \\
    \bottomrule
    \end{tabular}%
\end{table*}%

\begin{table*}[!h]
  \centering
  \footnotesize
  \setlength{\tabcolsep}{9pt} 
  \caption{Additional Ablation Study on the Base Model, Consistency Loss for Surface Normal Estimation. The \colorbox{first}{best} and \colorbox{second}{second-best} performances are highlighted.}
  \label{tab:ablation_normal_supp} 
    \begin{tabular}{llccccccccc}
    \toprule
    \multirow{2}{*}{\textbf{ID}} & \multirow{2}{*}{\textbf{Base Model}} & \multirow{2}{*}{\textbf{$\mathcal{L}_{Cons}$}} & \multicolumn{2}{c}{\textbf{NYUv2}} & \multicolumn{2}{c}{\textbf{Scannet}} & \multicolumn{2}{c}{\textbf{iBims-1}} & \multicolumn{2}{c}{\textbf{DIODE}}\\
    & & & \textbf{Mean $\downarrow$} & \textbf{11.25° $\uparrow$} & \textbf{Mean $\downarrow$} & \textbf{11.25° $\uparrow$} & \textbf{Mean $\downarrow$} & \textbf{11.25° $\uparrow$} & \textbf{Mean $\downarrow$} & \textbf{11.25° $\uparrow$}\\
    \midrule
        1 & Flux.1 &  & 16.6 & 57.7 & 15.0 & 62.1 & 17.0 & 65.1 & \cellcolor{second}19.9 & \cellcolor{first}44.7 \\ 
        2 & Flux.1 & \checkmark & 16.4 & 59.1 & 14.9 & 63.3 & 16.8 & 66.2 & \cellcolor{second}19.9 & 40.1 \\ 
        3 & Flux.1 Kontext &  & \cellcolor{second}15.8 & \cellcolor{second}60.0 & \cellcolor{second}14.2 & \cellcolor{second}65.2 & \cellcolor{second}15.8 & \cellcolor{second}68.6 & 20.1 & 42.0 \\ 
        4 & Flux.1 Kontext & \checkmark & \cellcolor{first}15.7 & \cellcolor{first}61.6 & \cellcolor{first}14.1 & \cellcolor{first}66.3 & \cellcolor{first}15.1 & \cellcolor{first}70.9 & \cellcolor{first}18.7 & \cellcolor{second}44.3 \\ 
    \bottomrule
    \end{tabular}%
\end{table*}%

\begin{table*}[!h]
\centering
\footnotesize
\setlength{\tabcolsep}{2.0pt}
\caption{Additional Ablation Study on the Base Model, Consistency Loss for Interactive Matting. The \colorbox{first}{best} and \colorbox{second}{second-best} performances are highlighted.}
\label{tab:ablation_matting_supp}
\begin{tabular}{llc *{5}{c} *{5}{c} *{5}{c}} 
        \toprule
            \multirow{2}{*}{\textbf{ID}} & \multirow{2}{*}{\textbf{Base Model}} & \multirow{2}{*}{\textbf{$\mathcal{L}_{Cons}$}} & \multicolumn{5}{c}{\textbf{AIM-500}} & \multicolumn{5}{c}{\textbf{P3M-500-NP}} & \multicolumn{5}{c}{\textbf{AM-2k}} \\
        \cmidrule(lr){4-8} \cmidrule(lr){9-13} \cmidrule(lr){14-18}  
        & & & \textbf{MSE$\downarrow$} & \textbf{MAD$\downarrow$} & \textbf{SAD$\downarrow$}  & \textbf{Grad$\downarrow$} & \textbf{Conn$\downarrow$} & \textbf{MSE$\downarrow$} & \textbf{MAD$\downarrow$} & \textbf{SAD$\downarrow$} & \textbf{Grad$\downarrow$} & \textbf{Conn$\downarrow$} & \textbf{MSE$\downarrow$} & \textbf{MAD$\downarrow$} & \textbf{SAD$\downarrow$} & \textbf{Grad$\downarrow$} & \textbf{Conn$\downarrow$} \\
        
        \midrule
        1 & FLUX.1 & ~ & 0.0495 & 0.085 & 144.25 & 23.74 & 72.02 & 0.0316 & 0.069 & 108.94 & 35.80 & 61.29 & 0.011 & 0.027 & 46.26 & 14.08 & 27.59 \\ 
        2 & FLUX.1 & \checkmark & 0.0490 & \cellcolor{second}0.084 & 142.23 & 23.54 & 70.31 & 0.0299 & \cellcolor{second}0.066 & 104.22 & 35.91 & 59.78 & 0.0102 & \cellcolor{second}0.024 & 45.13 & 13.58 & 25.87 \\ 
        3 & FLUX.1 Kontext & ~ & \cellcolor{second}0.0058 & \cellcolor{first}0.017 & \cellcolor{second}30.84 & \cellcolor{first}18.12 & \cellcolor{second}17.51 & \cellcolor{second}0.0034 & \cellcolor{first}0.011 & \cellcolor{second}22.64 & \cellcolor{first}10.81 & \cellcolor{second}12.85 & \cellcolor{second}0.0039 & \cellcolor{first}0.012 & \cellcolor{second}21.53 & \cellcolor{second}9.81 & \cellcolor{second}12.85 \\ 
        4 & FLUX.1 Kontext & \checkmark & \cellcolor{first}0.0057 & \cellcolor{first}0.017 & \cellcolor{first}29.14 &\cellcolor{second}18.28 & \cellcolor{first}15.73 & \cellcolor{first}0.0028 & \cellcolor{first}0.011 & \cellcolor{first}19.39 & \cellcolor{second}13.21 & \cellcolor{first}10.17 & \cellcolor{first}0.0037 & \cellcolor{first}0.012 & \cellcolor{first}20.42 & \cellcolor{first}9.61 & \cellcolor{first}9.94 \\ 
        \bottomrule
\end{tabular}

\end{table*}

\begin{table*}[!h]
  \centering
  \footnotesize
  \setlength{\tabcolsep}{8.5pt} 
  \caption{Additional Ablation Study on the Inference Steps of Monocular Depth Estimation task. The \colorbox{first}{best} and \colorbox{second}{second-best} performances are highlighted.}
  \label{tab:ablation_steps_depth_supp}
    \begin{tabular}{llcccccccccccc}
    \toprule
    \multirow{2}{*}{\textbf{Inference Steps}} & 
    \multicolumn{2}{c}{\textbf{NYUv2}} & \multicolumn{2}{c}{\textbf{KITTI}} & \multicolumn{2}{c}{\textbf{ETH3D}} & \multicolumn{2}{c}{\textbf{Scannet}} & \multicolumn{2}{c}{\textbf{DIODE}}\\
    & \textbf{AbsRel $\downarrow$} & \textbf{$\delta_1$ $\uparrow$} & \textbf{AbsRel $\downarrow$} & \textbf{$\delta_1$ $\uparrow$} & \textbf{AbsRel $\downarrow$} & \textbf{$\delta_1$ $\uparrow$} & \textbf{AbsRel $\downarrow$} & \textbf{$\delta_1$ $\uparrow$} & \textbf{AbsRel $\downarrow$} & \textbf{$\delta_1$ $\uparrow$}  \\
    \midrule
        1 & 4.425 & \cellcolor{second}97.616 & 7.913 & 94.472 & 4.283 & 98.256 & 4.886 & 97.334 & \cellcolor{second}24.848 & 81.429 \\ 
        2 & 4.423 & 97.614 & 7.910 & 94.470 & 4.281 & 98.254 & 4.888 & 97.333 & \cellcolor{first}24.846 & 81.436 \\ 
        4 & \cellcolor{first}4.198 & \cellcolor{first}97.644 & \cellcolor{first}7.375 & \cellcolor{second}95.185 & \cellcolor{second}3.637 & \cellcolor{second}98.494 & \cellcolor{second}4.602 & \cellcolor{second}97.425 & \cellcolor{first}24.846 & \cellcolor{first}81.589 \\ 
        10 & \cellcolor{second}4.264 & 97.555 & \cellcolor{second}7.417 & \cellcolor{first}95.220 & \cellcolor{first}3.631 & \cellcolor{first}98.517 & \cellcolor{first}4.592 & \cellcolor{first}97.433 & 25.026 & \cellcolor{second}81.438 \\ 
        25 & 4.312 & 97.475 & 7.487 & 95.134 & 3.669 & 98.482 & 4.615 & 97.411 & 25.145 & 81.365 \\
    \bottomrule
    \end{tabular}%
\end{table*}%

\begin{table*}[!h]
  \centering
  \footnotesize
  \setlength{\tabcolsep}{13pt} 
  \caption{Ablation on the Inference Steps for Surface Normal Estimation. The \colorbox{first}{best} and \colorbox{second}{second-best} performances are highlighted.}
  \label{tab:ablation_steps_normal_supp} 
    \begin{tabular}{llccccccccc}
    \toprule
    \multirow{2}{*}{\textbf{Inference Steps}} & \multicolumn{2}{c}{\textbf{NYUv2}} & \multicolumn{2}{c}{\textbf{Scannet}} & \multicolumn{2}{c}{\textbf{iBims-1}} & \multicolumn{2}{c}{\textbf{DIODE}}\\
    & \textbf{Mean $\downarrow$} & \textbf{11.25° $\uparrow$} & \textbf{Mean $\downarrow$} & \textbf{11.25° $\uparrow$} & \textbf{Mean $\downarrow$} & \textbf{11.25° $\uparrow$} & \textbf{Mean $\downarrow$} & \textbf{11.25° $\uparrow$}\\
    \midrule
        1 & \cellcolor{second}15.668 & 61.584 & \cellcolor{second}14.105 & 66.347 & 15.136 & 70.936 & 18.710 & \cellcolor{second}44.287 \\ 
        2 & \cellcolor{first}15.663 & 61.619 & \cellcolor{first}14.101 & 66.365 & \cellcolor{second}15.119 & 70.991 & 18.683 & \cellcolor{first}44.389 \\ 
        4 & 16.239 & \cellcolor{first}62.134 & 14.254 & \cellcolor{first}67.503 & \cellcolor{first}14.923 & \cellcolor{first}72.063 & \cellcolor{first}18.507 & 44.019 \\ 
        10 & 16.627 & \cellcolor{second}61.761 & 14.599 & \cellcolor{second}67.270 & 15.183 & \cellcolor{second}71.938 & \cellcolor{second}18.628 & 44.079 \\ 
        25 & 16.830 & 61.445 & 14.752 & 67.035 & 15.344 & 71.746 & 18.710 & \cellcolor{second}44.287 \\
    \bottomrule
    \end{tabular}%
\end{table*}%

 \begin{table*}[!h]
\centering
\footnotesize
\setlength{\tabcolsep}{3.3pt}
\caption{Additional Ablation Study on the Base Model, Consistency Loss for Interactive Matting. The \colorbox{first}{best} and \colorbox{second}{second-best} performances are highlighted.}
\label{tab:ablation_steps_matting_supp}
\begin{tabular}{l *{5}{c} *{5}{c} *{5}{c}} 
        \toprule
         \multirow{2}{*}{\textbf{Inference Steps}} & \multicolumn{5}{c}{\textbf{AIM-500}} & \multicolumn{5}{c}{\textbf{P3M-500-NP}} & \multicolumn{5}{c}{\textbf{AM-2k}} \\
        \cmidrule(lr){2-6} \cmidrule(lr){7-11} \cmidrule(lr){12-16}  
        & \textbf{MSE$\downarrow$} & \textbf{MAD$\downarrow$} & \textbf{SAD$\downarrow$}  & \textbf{Grad$\downarrow$} & \textbf{Conn$\downarrow$} & \textbf{MSE$\downarrow$} & \textbf{MAD$\downarrow$} & \textbf{SAD$\downarrow$} & \textbf{Grad$\downarrow$} & \textbf{Conn$\downarrow$} & \textbf{MSE$\downarrow$} & \textbf{MAD$\downarrow$} & \textbf{SAD$\downarrow$} & \textbf{Grad$\downarrow$} & \textbf{Conn$\downarrow$} \\
        
        \midrule
         1 & 0.0057 & 0.017 & 29.14 & 18.28 & 15.73 & 0.0028 & 0.011 & 19.39 & 13.21 & 10.17 & 0.0037 & 0.012 & 20.42 & 9.61 & 9.94 \\ 
        2 & \cellcolor{second}0.0056 & 0.017 & 28.32 & 18.16 & 15.35 & 0.0028 & 0.011 & 19.13 & 13.01 & 10.06 & 0.0034 & 0.012 & 20.16 & 9.59 & 9.93 \\ 
        4 & \cellcolor{first}0.0055 & \cellcolor{first}0.013 & \cellcolor{second}21.97 & \cellcolor{first}16.14 & 13.66 & \cellcolor{first}0.0022 & \cellcolor{first}0.006 & \cellcolor{second}11.14 & \cellcolor{first}12.17 & 8.04 & \cellcolor{first}0.0032 & \cellcolor{first}0.008 & \cellcolor{second}12.74 & \cellcolor{first}8.52 & \cellcolor{second}7.99 \\ 
        10 & 0.0059 & \cellcolor{first}0.013 & \cellcolor{first}21.74 & \cellcolor{second}16.59 & \cellcolor{second}13.34 & \cellcolor{second}0.0025 & \cellcolor{first}0.006 & \cellcolor{first}10.92 & \cellcolor{second}12.54 & \cellcolor{first}7.63 & \cellcolor{second}0.0034 & \cellcolor{first}0.008 & \cellcolor{first}12.66 & \cellcolor{second}8.95 & \cellcolor{first}7.96 \\ 
        25 & 0.0059 & \cellcolor{second}0.014 & 23.48 & 17.32 & \cellcolor{first}13.20 & 0.0029 & \cellcolor{second}0.008 & 13.37 & 13.25 & \cellcolor{second}7.65 & \cellcolor{second}0.0034 & \cellcolor{second}0.009 & 14.74 & 9.59 & 8.10 \\ 
        \bottomrule
\end{tabular}

\end{table*}

\section{Additional Qualitive Results}
\subsection{Comparison of other SOTA models}
\begin{figure*}[!p]
    \centering
    \includegraphics[width=\linewidth]{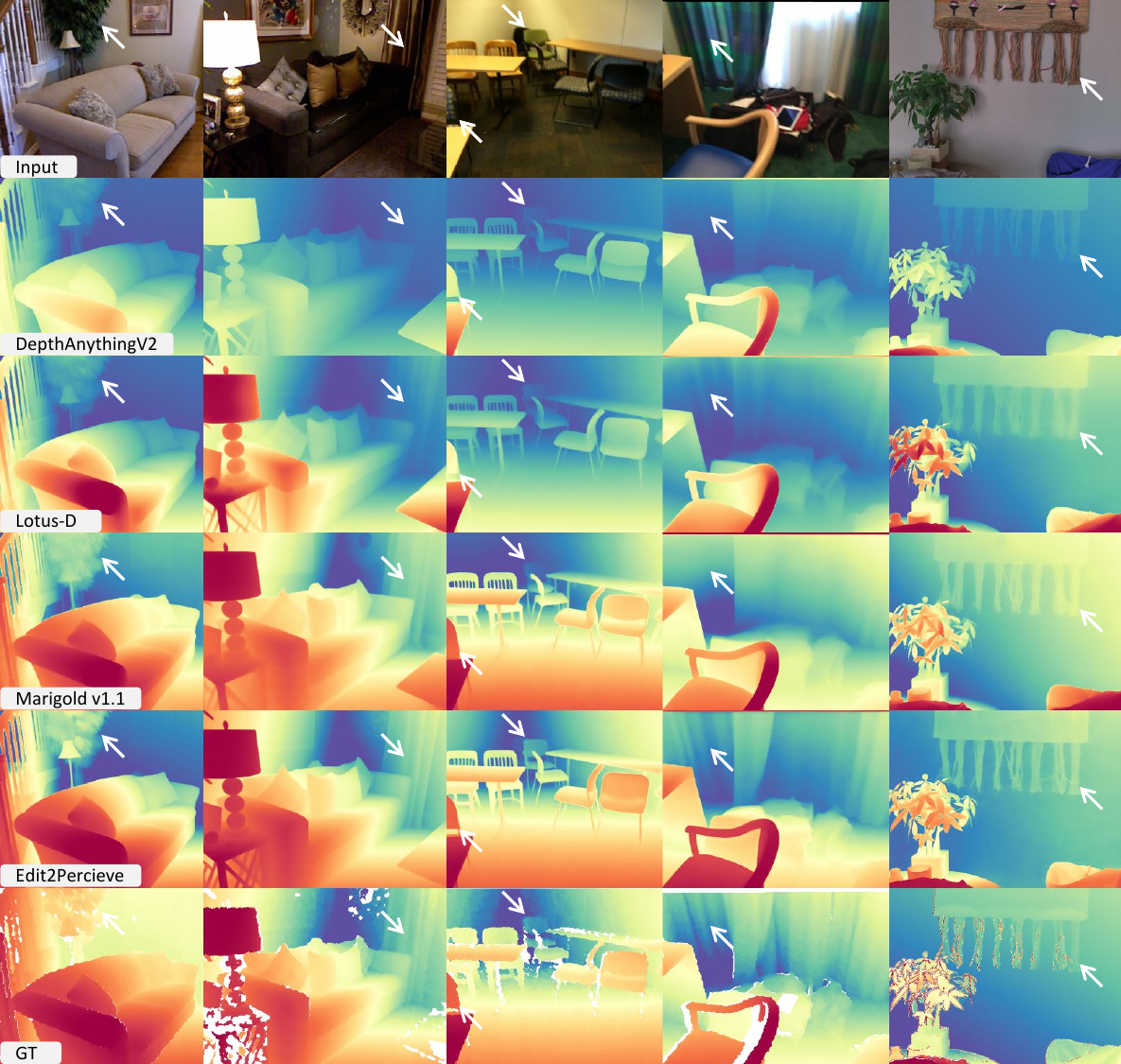}
    \caption{\textbf{Additional Qualitative Comparisons for Zero-Shot Monocular Depth Estimation.} Our method consistently produces more detailed and structurally coherent depth maps compared to other state-of-the-art methods across a variety of challenging indoor and outdoor scenes.}
    \label{fig:depth_supp}
\end{figure*}

Figure~\ref{fig:depth_supp} provides additional qualitative comparisons for zero-shot monocular depth estimation. We observe that our model, Edit2Perceive, demonstrates superior performance in capturing complex scene geometry compared to prior works. For instance, our method accurately reconstructs fine-grained details such as the folds of the curtains (second and fourth columns) and the intricate structure of pine needles within shadowed regions (first column), highlighting its powerful capability for detailed geometric reasoning.

\begin{figure*}[!p]
    \centering
    \includegraphics[width=\linewidth]{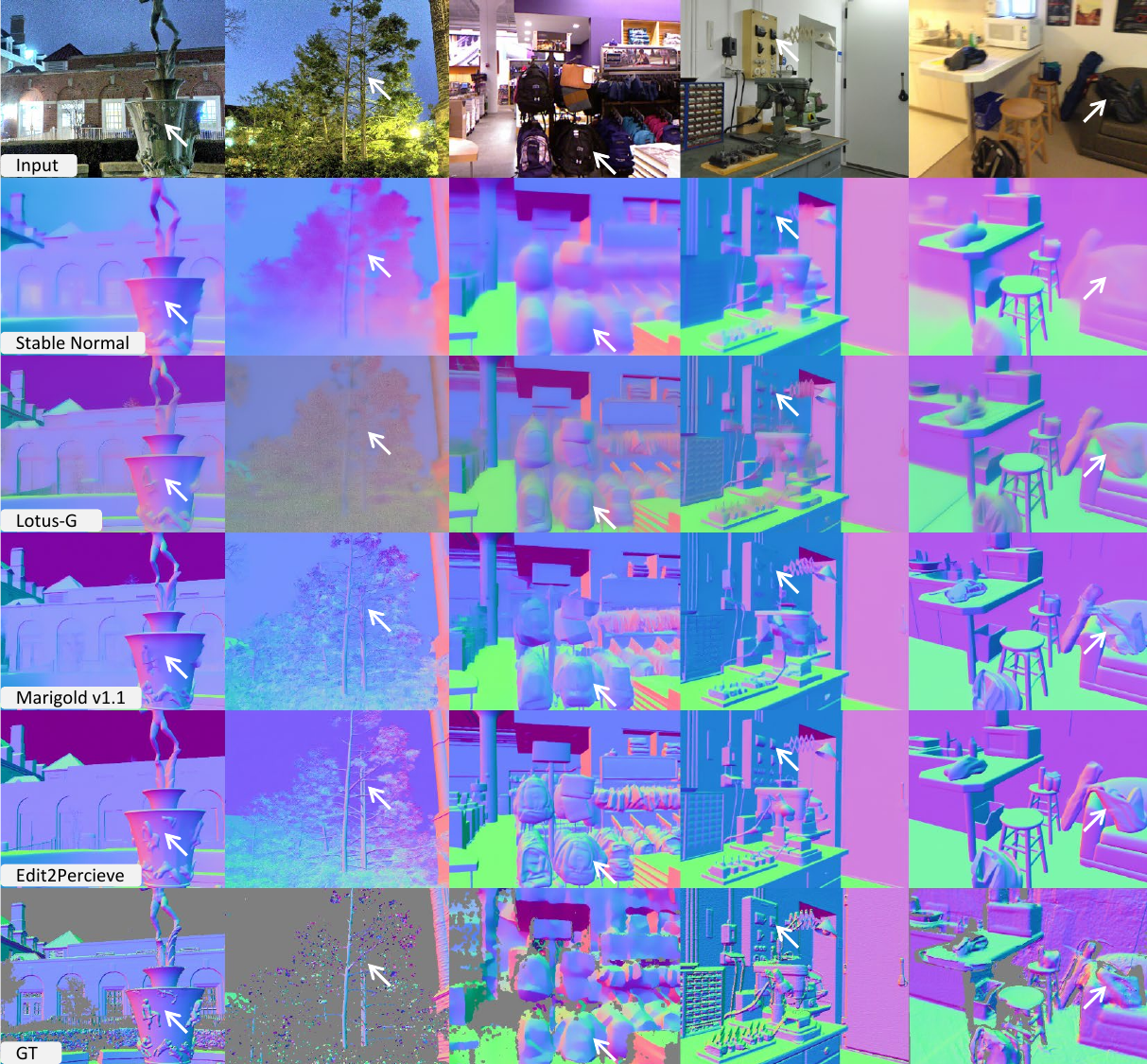}
    \caption{\textbf{Additional Qualitative Comparisons for Zero-Shot Surface Normal Estimation.} Compared to other methods, our model demonstrates a superior ability to capture fine-grained surface details and subtle curvatures, such as the texture of tree bark (second column) and fabric patterns (third column).}
    \label{fig:normal_supp}
\end{figure*}

In Figure~\ref{fig:normal_supp}, we present further qualitative comparisons for zero-shot surface normal estimation. Our model excels in scenarios with complex and subtle textures. Notably, it successfully captures the rough texture of the tree bark and the delicate structure of leaves (second column), as well as the fine surface patterns on the backpack (third column). This demonstrates the model's robustness in recovering detailed surface geometry from challenging in-the-wild images.

\begin{figure*}[!p]
    \centering
    \includegraphics[width=\linewidth]{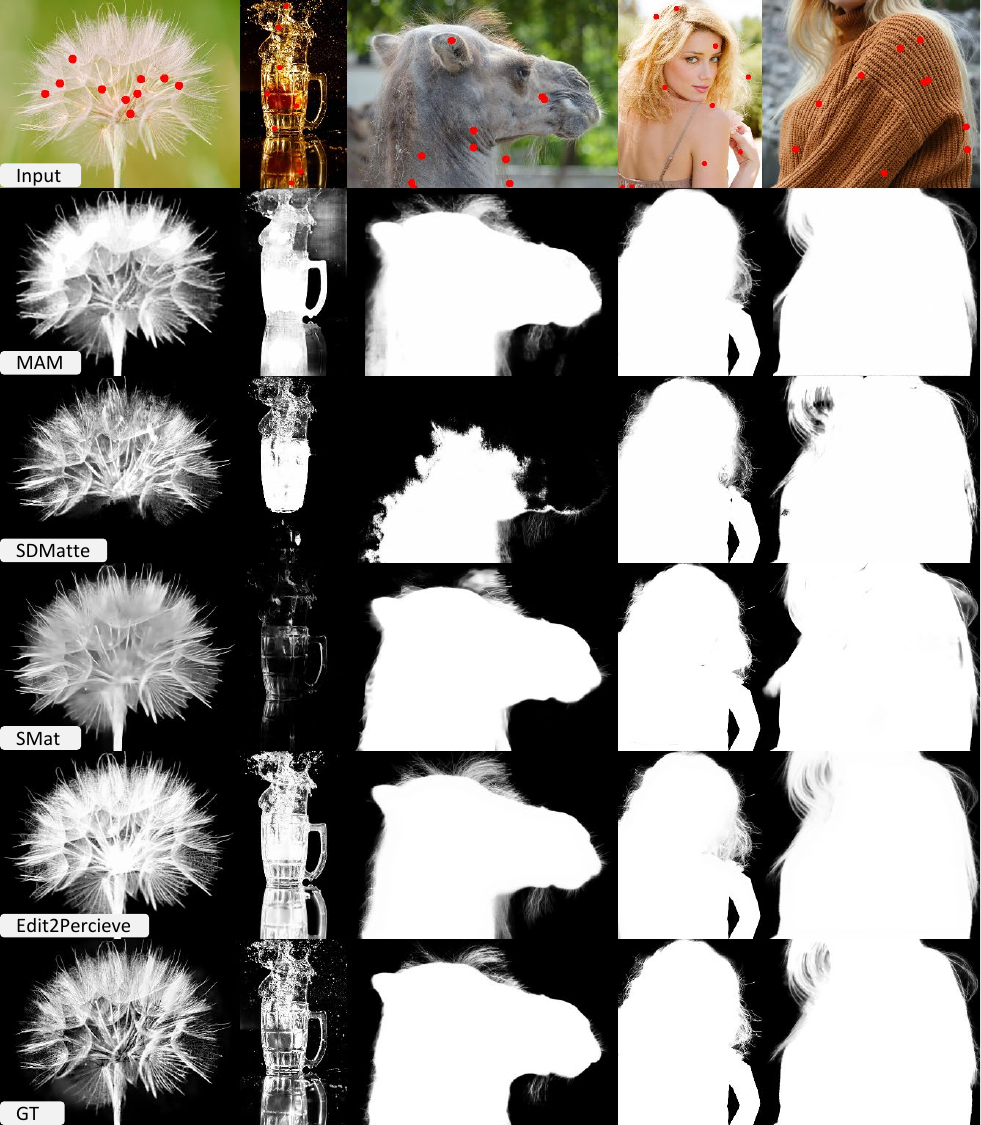}
    \caption{\textbf{Additional Qualitative Comparisons for Interactive Matting.} Our method excels at handling challenging cases, including extremely fine structures like hair and feathers, as well as semi-transparent materials like glass and water droplets, producing significantly cleaner and more accurate alpha mattes.}
    \label{fig:matting_supp}
\end{figure*}

Figure~\ref{fig:matting_supp} illustrates the superior performance of our model on the interactive matting task. Edit2Perceive exhibits exceptional capability in handling extremely fine details and challenging materials. It accurately delineates delicate structures like feathers and hair, and correctly handles semi-transparent objects such as glass cups and water droplets, setting it apart from competing methods.

\subsection{Visual Abalation Study of Components}
\begin{figure*}
    \centering
    \includegraphics[width=\linewidth]{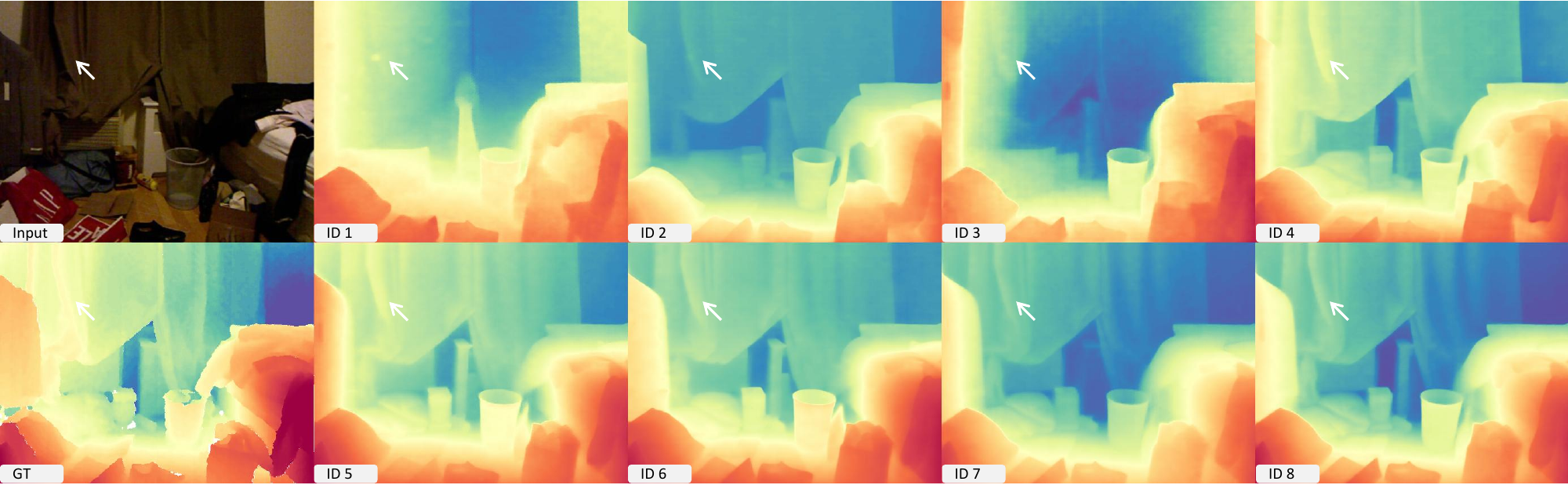}
    \caption{\textbf{Visual Ablation Study for Monocular Depth Estimation.} Each row ID  are from Table~\ref{tab:ablation_depth_supp}, allowing for a direct visual assessment of each component's impact. These results visually confirm the quantitative findings: the I2I base model, the consistency loss, and our Sqrt depth mapping each contribute significantly to the final performance.}
    \label{fig:depth_ids}
\end{figure*}

To visually dissect the contribution of each component, we present the ablation results for depth estimation in Figure~\ref{fig:depth_ids}.
\textbf{Base Model:} Comparing models with identical settings but different base models (e.g., ID 1 vs. 5, ID 2 vs. 6, etc.), it is evident that the FLUX.1 Kontext (I2I) based models (IDs 5-8) consistently produce more accurate and structurally sound results than their FLUX.1 (T2I) counterparts (IDs 1-4).
\textbf{Consistency Loss:} The effect of our pixel-space consistency loss can be seen by comparing adjacent columns (e.g., ID 1 vs. 2, ID 5 vs. 6). The addition of $\mathcal{L}_{\text{Cons}}$ (IDs 2,4,6,8) consistently enhances fine-grained details, as highlighted by the sharper reconstruction of the curtains (indicated by arrows).
\textbf{Depth Mapping:} Comparing different depth normalization methods (e.g., ID 1 vs. 3, ID 2 vs. 4), our proposed Sqrt mapping (IDs 3,4,7,8) yields visibly superior results compared to the Uniform mapping (IDs 1,2,5,6), particularly in preserving depth variations.

\begin{figure*}
    \centering
    \includegraphics[width=\linewidth]{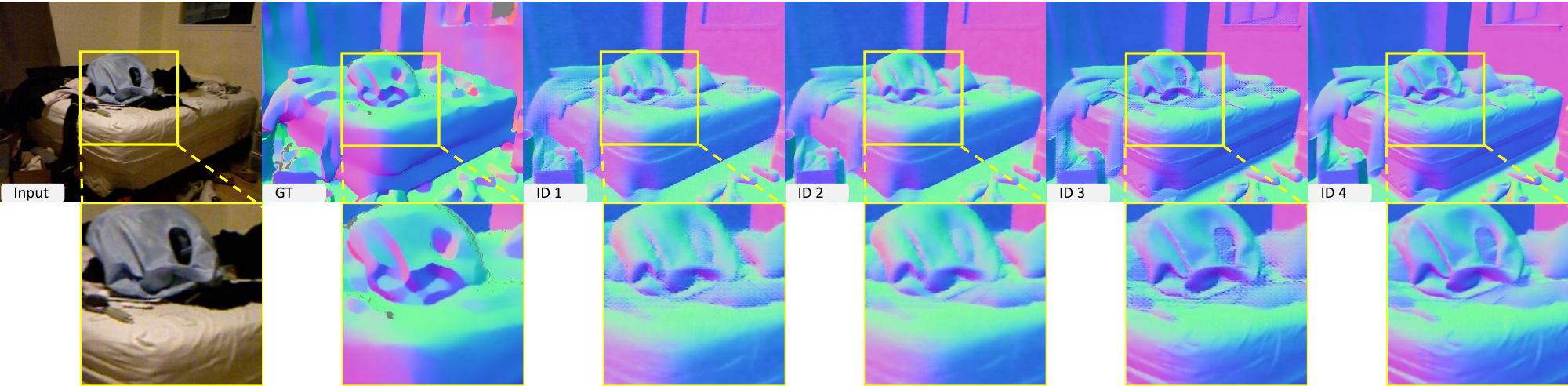}
    \caption{\textbf{Visual Ablation Study for Surface Normal Estimation.} Each column corresponds to an ID from Table~\ref{tab:ablation_normal_supp}. The zoomed-in regions (below) highlight how our consistency loss effectively removes speckled artifacts (ID 1 vs. 2 and 3 vs. 4) and how the I2I base model better captures complex geometry (ID 1-2 vs. 3-4).}
    \label{fig:normal_ids}
\end{figure*}

Figure~\ref{fig:normal_ids} visualizes the ablation study for surface normal estimation. We observe that without the consistency loss (IDs 1 \& 3), the predictions are prone to speckled artifacts and noisy patterns. The introduction of our pixel-space supervision, $\mathcal{L}_{\text{Cons}}$, (IDs 2 \& 4) significantly mitigates these issues, resulting in much smoother and more coherent normal maps. Furthermore, comparing the base models, FLUX.1 Kontext (IDs 3 \& 4) demonstrates a markedly improved ability to discern complex edges compared to FLUX.1 (IDs 1 \& 2).

\begin{figure*}
    \centering
    \includegraphics[width=\linewidth]{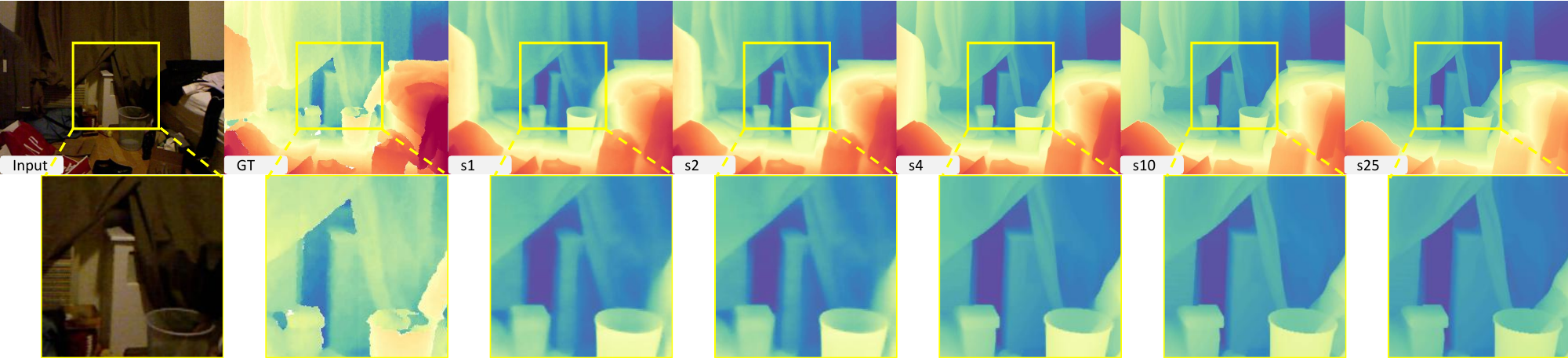}
    \caption{\textbf{Visualizing the Effect of Inference Steps on Depth Estimation.} The zoomed-in regions (below) show that while increasing the number of steps from 1 to 4 offers slight improvements in detail, further steps yield diminishing returns. This demonstrates that our single-step inference already achieves high-quality results.}
    \label{fig:depth_steps}
\end{figure*}

\begin{figure*}
    \centering
    \includegraphics[width=\linewidth]{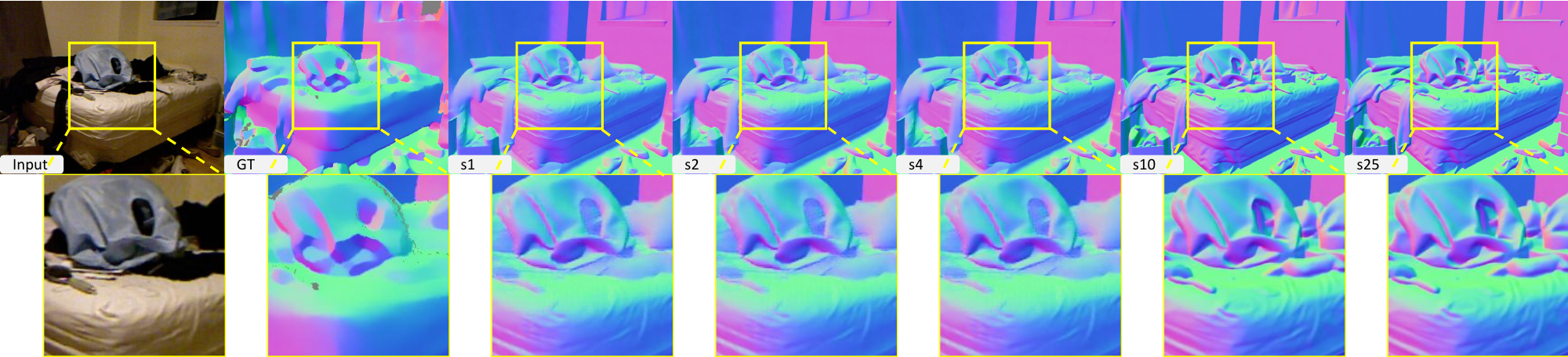}
    \caption{\textbf{Visualizing the Effect of Inference Steps on Normal Estimation.} Similar to depth estimation, we observe that single-step inference produces results comparable to multi-step inference. The performance gain from additional steps is marginal, highlighting the efficiency of our method.}
    \label{fig:normal_steps}
\end{figure*}

Figures~\ref{fig:depth_steps} and~\ref{fig:normal_steps} visualize the effect of varying the number of inference steps for depth and normal estimation, respectively. We observe that while additional steps can offer marginal refinements in edge sharpness, our single-step inference already produces high-quality and structurally coherent results. The performance gain from multi-step inference is minimal, confirming that our approach offers an excellent trade-off between efficiency and quality with negligible and acceptable performance loss.

\clearpage

\clearpage

\clearpage

\end{document}